\documentclass[3p,times]{elsarticle}

\usepackage{ecrc}
 \usepackage{makecell}

\volume{00}

\firstpage{1}

\journalname{Communications in Nonlinear Science
and Numerical Simulation}

\runauth{A.N. Hasmi, H. Hatzikirou,  H. Susanto}

\jid{JCP}

\jnltitlelogo{CNSNS}

\CopyrightLine{2026}{Published by Elsevier Ltd.}

\usepackage{amssymb}
\usepackage{xcolor} 

\usepackage{amsmath}

\usepackage[figuresright]{rotating}

\usepackage{verbatim}
\usepackage{amsmath}
\usepackage{amsthm}
\usepackage{amssymb}
\usepackage{hyperref}
\usepackage{microtype}
\usepackage{multirow}

\begin{document}

\begin{frontmatter}



\dochead{}

\title{
Phase space integrity in neural network models of Hamiltonian dynamics: A Lagrangian descriptor approach
}
\author{Abrari Noor Hasmi\fnref{label1,label2}}
\ead{100060615@ku.ac.ae;abrari.hasmi@ku.ac.ae}
\author{Haralampos Hatzikirou\fnref{label1}}
\ead{haralampos.hatzikirou@ku.ac.ae}
\author{Hadi Susanto\corref{cor1}\fnref{label1}}
\ead{hadi.susanto@yandex.com}
\fnref{label1}
\fntext[label1]{Department of Mathematics, Khalifa University of Science and Technology, PO Box 127788, Abu Dhabi, United Arab Emirates}

\cortext[cor1]{Corresponding Author}

\fntext[label2]{Public Health and Epidemeology Department, Khalifa University of Science and Technology, PO Box 127788, Abu Dhabi, United Arab Emirates.}

\date{\today}
\begin{abstract}
We propose Lagrangian Descriptors (LDs) as a diagnostic framework for evaluating neural network models of Hamiltonian systems beyond conventional trajectory-based metrics. Standard error measures quantify short-term predictive accuracy but provide little insight into global geometric structures such as orbits and separatrices. Existing evaluation tools in dissipative systems are inadequate for Hamiltonian dynamics due to fundamental differences in the systems. By constructing probability density functions weighted by LD values, we embed geometric information into a statistical framework suitable for information-theoretic comparison. We benchmark physically constrained architectures (SympNet, H\'enonNet, Generalized Hamiltonian Neural Networks) against data-driven Reservoir Computing across two canonical systems. For the Duffing oscillator, all models recover the homoclinic orbit geometry with modest data requirements, though their accuracy near critical structures varies. For the three-mode nonlinear Schr\"odinger equation, however, clear differences emerge: symplectic architectures preserve energy but distort phase-space topology, while Reservoir Computing, despite lacking explicit physical constraints, reproduces the homoclinic structure with high fidelity. These results demonstrate the value of LD-based diagnostics for assessing not only predictive performance but also the global dynamical integrity of learned Hamiltonian models.
\end{abstract}


\end{frontmatter}

\section{Introduction\label{sec:level1}}
The intersection of artificial intelligence (AI) and the natural sciences has advanced rapidly in recent years, particularly in the modeling of physical dynamics. Modern AI-based approaches, such as Physics-Informed Neural Networks (PINNs) \cite{raissi2019PhysicsinformedNeural}, Sparse Identification of Nonlinear Dynamics (SINDy) \cite{Brunton2016}, Neural Ordinary Differential Equations (Neural ODEs) \cite{chen2018NeuralOrdinarya}, and Fourier Neural Operators (FNOs) \cite{li2021FourierNeurala}, have demonstrated remarkable success in capturing complex systems. These models, however, only capture a subset of the vast landscape of dynamical systems.

While AI-driven models are highly effective for dissipative systems, their extension to Hamiltonian systems introduces unique challenges \cite{greydanus2019HamiltonianNeural}. Dissipative systems typically exhibit trajectories converging to low-dimensional attractors, which simplifies the learning task. In contrast, Hamiltonian systems conserve energy, precluding attractors and allowing trajectories to evolve on high-dimensional phase spaces. These trajectories can be either regular (confined to invariant tori in integrable systems) or chaotic (characterized by sensitive dependence on initial conditions and complex phase-space exploration, which may—but need not—be ergodic within invariant subsets). \cite{bountis2012ComplexHamiltonian}. The coexistence of ordered and chaotic dynamics, together with the absence of dissipative convergence, creates a particularly demanding setting for machine learning models.

To address these challenges, recent studies have introduced neural architectures that explicitly preserve Hamiltonian structure. Notable examples include Hamiltonian Neural Networks (HNNs) \cite{greydanus2019HamiltonianNeural}, which learn the Hamiltonian function directly, and Symplectic Recurrent Neural Networks (SRNNs) \cite{Chen2020Symplectic}, which embed symplectic integration within recurrent units. More recently, Horn et al.\ \cite{horn2025GeneralizedFramework} proposed the Generalized Hamiltonian Neural Network (GHNN) framework, which unifies and extends earlier efforts by combining neural representations of Hamiltonian components with symplectic integration schemes. This architecture generalizes and combines several prior symplectic models, including SympNets \cite{jin2020SympNetsIntrinsic} and H\'enonNets \cite{burby2020FastNeural}, offering a comprehensive structure-preserving framework. Interestingly, even methods not explicitly designed for Hamiltonian dynamics, such as Reservoir Computing (RC) \cite{zhangLearningHamiltonianDynamics2021}, have achieved notable predictive accuracy, though they often struggle to preserve geometric invariants over long time horizons.

Despite these advances, most evaluation studies rely primarily on trajectory accuracy, emphasizing short-term prediction. Such metrics, however, fail to assess whether models capture the global phase-space geometry, including invariant manifolds and homoclinic tangles, which govern long-term behavior. Traditional tools like Lyapunov Exponents (LEs) are widely used in dissipative settings \cite{pathak_using_2017,kobayashi_dynamical_2021,kobayashi2024LyapunovAnalysis,gallicchio_design_2018,Vlachas2020}, but are less practical for Hamiltonian systems due to their sensitivity to initial conditions and the absence of attracting structures \cite{strogatz2015NonlinearDynamics}. Similarly, Poincar\'e sections are powerful for visualizing low-dimensional systems (e.g., \cite{zhangLearningHamiltonianDynamics2021} in RC), but scale poorly to higher dimensions and depend heavily on the chosen section.

A growing body of research has sought to move beyond short-term forecasting, focusing instead on global structure and statistical properties of learned models, especially for dissipative systems. For instance, studies have compared Lyapunov spectra \cite{pathak_using_2017,kobayashi_dynamical_2021,gallicchio_design_2018}, reconstructed attractors \cite{Kim2021}, and probability distributions of physical observables \cite{Chattopadhyay2020,Huhn2021} between models and ground truth systems. Yet, such approaches remain underexplored for Hamiltonian dynamics.

Lagrangian Descriptors (LDs) provide a promising alternative \cite{mendoza2010HiddenGeometry,mancho2013LagrangianDescriptors}. By integrating trajectory-dependent quantities over finite times, LDs reveal invariant manifolds, homoclinic tangles, and other global geometric structures with modest computational effort. Unlike LEs, which characterize asymptotic divergence, LDs emphasize finite-time dynamics and provide a global “climate” of the phase space, making them particularly suited for evaluating data-driven Hamiltonian models. This study extends our preliminary findings \cite{hasmi2025PhaseSpace} by presenting the first comprehensive application of LDs for the assessment of neural models of Hamiltonian dynamics.

The contributions of this work are twofold. First, we establish a framework that complements conventional error metrics with LD-based analysis, enabling systematic evaluation of how well models reproduce global phase-space structures. Second, we benchmark several architectures—physically constrained networks (SympNet, H\'enonNet, GHNN) and RC—on Hamiltonian systems of varying complexity. To our knowledge, this is the first systematic study to evaluate neural network (NN) models of Hamiltonian dynamics using a geometry-sensitive LD-based framework.

Our focus is on Hamiltonian systems with homoclinic structures, which are central to the onset of chaos and global topology of phase space. Homoclinic orbits, connecting stable and unstable manifolds of a saddle point, play a key role in organizing dynamics. We consider two paradigmatic examples: the Duffing oscillator and a reduced-order model of the nonlinear Schr\"odinger equation (NLSE), both of which exhibit rich dynamics including homoclinic bifurcations and chaos. These provide stringent benchmarks for testing data-driven models.

Our methodology compares LD fields derived from ground truth solutions against those generated by neural models. We then construct probability density functions (PDFs) weighted by LD values, translating geometric structures into a statistical framework. Model performance is quantified using information-theoretic measures such as the Kullback--Leibler (KL) divergence. This approach not only evaluates the accuracy of global structures but also prioritizes regions of dynamical significance, offering both diagnostic and computational advantages.

\section{Lagrangian Descriptors (LDs)\label{sec:LD}}
\subsection{Definition}

LDs, first introduced in \cite{mendoza2010HiddenGeometry}, were originally proposed to uncover invariant manifolds in dynamical systems. Since then, the method has been extensively developed and applied in various contexts, including the visualization of hyperbolic invariant manifolds in high-dimensional systems \cite{naik2019FindingNHIM}, the identification of chaotic regions \cite{revuelta2019UnveilingChaotic}, and as an efficient indicator of chaos \cite{hillebrand2022QuantifyingChaos}. LDs have become a powerful diagnostic tool due to their conceptual simplicity and straightforward numerical implementation. 

Formally, consider an autonomous Hamiltonian system with state space $U \subset \mathbb{R}^{2n}$ and evolution operator $\Phi : I \times U \to U$, where $I \subset \mathbb{R}$ is the time domain of definition. The operator $\Phi$ satisfies:
\begin{itemize}
    \item Initial condition: $\Phi(0,u) = u$ for all $u \in U$,
    \item Group property: $\Phi(t+s,u) = \Phi(t,\Phi(s,u))$ for all $t,s \in I$ and $u \in U$. 
\end{itemize}

The trajectory starting from $u_{0}$ is given by $\gamma_{u_{0}}(t) = \Phi(t,u_{0})$. The LD is defined as a scalar field $LD : U \to \mathbb{R}$ over a finite integration interval $I_0 \subset I$:
\begin{equation}
    LD(u_{0}) = \int_{I_0} \mathcal{M}(\gamma_{u_{0}}(t))\,dt , \label{eq:LD_definition}
\end{equation}
where $\mathcal{M}$ is a positive-definite functional and $I_0 = [t_0-\tau,\, t_0+\tau]$. A widely used choice for $\mathcal{M}$ is based on the trajectory arc length \cite{mancho2013LagrangianDescriptors,lopesino2017TheoreticalFramework}:
\begin{equation}
    \mathcal{M}(\gamma) = |\dot{q}|^c + |\dot{p}|^c, \qquad 0<c<1, \label{eq:LD_arc_length}
\end{equation}
where $q$ and $p$ denote canonical coordinates. The exponent $c \in (0,1)$ ensures that stable and unstable manifolds appear as singular features in the LD field. To emphasize the dependence on the underlying evolution operator, we write $LD_\Phi$ when appropriate.

For autonomous Hamiltonian systems, time-translation invariance implies that the choice of initial integration time $t_0$ is immaterial. We therefore adopt the symmetric interval $I_0 = [-\tau,\, \tau]$. In this case, the LD naturally decomposes into forward and backward components:
\begin{equation}
    LD(u_{0}) = LD_F(u_{0}) + LD_B(u_{0}),
\end{equation}
\[
    LD_F(u_{0}) = \int_{0}^{\tau} \mathcal{M}(\gamma_{u_{0}}(t))\,dt, \qquad 
    LD_B(u_{0}) = \int_{-\tau}^{0} \mathcal{M}(\gamma_{u_{0}}(t))\,dt.
\]

In the case of purely elliptic regions of the phase space, LD values are smooth, and the isosurface of LD correspond to the isoenergy surface (see \ref{App:LD_Ham} and \cite{lopesino2017TheoreticalFramework}). While for hyperbolic regions, the LD field exhibits singular features that align with stable and unstable manifolds \cite{lopesino2017TheoreticalFramework}. 

For initial conditions $u_0$ corresponding to regular (non-chaotic) motion, the normalized descriptor
\begin{equation}
    \hat{LD}(u_0) = \frac{LD(u_0)}{\tau},
\end{equation}
converges to a constant asymptotic value as $\tau \to \infty$ \cite{montes2021LagrangianDescriptors}. This convergence implies that, beyond a certain threshold, the precise choice of $\tau$ has little effect on discriminatory power. This property also motivates our PDF formulation (introduced in the next section), which provides an alternative normalization scheme for LDs.

In practice, LD computation requires both forward and backward trajectory integration. Backward integration can be computationally expensive, especially in nonlocal or memory-dependent systems. For example, in fractional differential equations \cite{theron2025UsingLagrangian}, backward integration requires iterative nonlinear solvers such as Newton–Raphson, which scale poorly with system dimension and integration time. In contrast, symplectic neural architectures (SympNet, H\'enonNet, GHNN) avoid this difficulty: their exact invertibility guarantees that backward trajectories can be obtained analytically with the same computational cost as forward propagation \cite{jin2020SympNetsIntrinsic}. For RC, backward propagation is not inherently supported due to its recurrent design. However, by exploiting the fixed, untrained reservoir, a backward model can be constructed by training the readout layer on time-reversed sequences while keeping the reservoir matrix fixed, enabling efficient approximation of backward dynamics.

A key advantage of LDs is that they require only trajectory data, without explicit knowledge of the governing equations or Hamiltonian. As scalar fields encoding global phase-space structure, LDs provide a direct means of detecting invariant manifolds, which manifest as singular features in the LD field \cite{mancho2013LagrangianDescriptors,lopesino2017TheoreticalFramework}.
The computation of LDs involves integrating a positive-definite functional along trajectories over a finite time interval $[-\tau, \tau]$. For a grid of $N_g$ initial conditions with $N_t$ time steps, LD evaluation requires both forward and backward trajectory integrations, effectively doubling the computational cost compared to forward-only simulation. The total computational cost scales as $\mathcal{O}(N_g \times N_t \times C_{\text{model}})$, where $C_{\text{model}}$ represents the per-step computational cost of the underlying dynamical model. For numerical integration of the reference system, $C_{\text{model}}$ corresponds to the cost of evaluating the vector field and performing one integration step. For NN models, it reflects the inference cost. The computational time can be substantially reduced through parallelization, as trajectory integrations from different initial conditions are completely independent.

\subsection{Probabilistic Point of View of LDs}

Direct comparison of evolution operators $\Phi$, which belong to an infinite-dimensional function space, poses significant computational and theoretical challenges. To overcome this, we treat LDs as dynamical observables that encode essential geometric features of $\Phi$. A related perspective was recently presented in \cite{chen2024LagrangianDescriptors}, where the authors examined uncertainty in the distribution of state variables. 

By interpreting the state space $U$ as a probability space, we construct a weighted PDF over $U$:
\begin{equation}\label{eq:PDF_LD}
\rho_{U}(u) = \frac{g(LD_{\Phi}(u))}{Z}, 
\qquad 
Z = \int_{U} g(LD_{\Phi}(u))\,du,
\end{equation}
where $g:\mathbb{R}^+ \to \mathbb{R}^+$ is a monotonic weighting function determining how LD values influence the probability density, and $\mathbb{R}^+$ denotes the set of non-negative real numbers. Possible choices include:
\begin{itemize}
\item Power law: $g(x) = x^m$, which emphasizes regions according to the magnitude of the LD values;
\item Exponential family: $g(x) = e^{-\beta x}$, inspired by statistical mechanics where the LD plays a role analogous to a Hamiltonian or energy function.
\end{itemize}

This probabilistic formulation of LDs offers several advantages:
\begin{itemize}
\item \textbf{Geometric interpretation:} The PDF $\rho_U$ captures the geometric organization of the phase space, with LD values serving as weights that highlight dynamically significant regions such as invariant manifolds and homoclinic orbits.
\item \textbf{Efficient exploration:} By assigning greater weight to regions of dynamical importance, the PDF enhances exploration of the phase space, revealing critical structures that may be obscured by conventional metrics.
\end{itemize}

For computational simplicity, we discretize the phase space using a uniform grid, which suffices for the low-dimensional systems studied here. For higher-dimensional systems, adaptive approaches such as importance sampling or Markov Chain Monte Carlo would become essential, focusing computational effort on dynamically relevant regions (e.g., near homoclinic orbits or saddle points) while preserving statistical accuracy. This extension is left for future work.

To compare phase-space distributions between the reference dynamics $\rho$ and the model predictions $\hat{\rho}$, we employ the KL divergence \cite{kullback1951information}:
\begin{equation}
    D_{KL}(\rho \,\|\, \hat{\rho}) 
    = \int_{U} \rho(u)\, \log \frac{\rho(u)}{\hat{\rho}(u)}\,du.
\end{equation}
This information-theoretic quantity measures the expected logarithmic difference between the two distributions, with $D_{KL}(\rho \,\|\, \hat{\rho}) = 0$ if and only if $\rho = \hat{\rho}$ almost everywhere. Although not a true metric---since it is asymmetric and does not satisfy the triangle inequality---the KL divergence is well suited to our application because it emphasizes regions of high probability in the reference distribution, i.e., precisely those regions where accurate reconstruction of dynamical structures is most critical.

In this setting, $\rho$ can be interpreted as the likelihood of observing particular states in phase space, where the LD weighting highlights hyperbolic structures such as stable and unstable manifolds. The idea of comparing probability distributions of data-driven models has been used in related contexts, for example, in~\cite{Huhn2021}, where the authors compared PDFs near attractors of an acoustic resonator.

As previously discussed, the PDF formulation provides a natural normalization for LDs. As an illustration of this point, consider the case of the harmonic oscillator, where the LD can be computed analytically (see \ref{App:LD_Ham}). In this case, $\rho_U \sim g(H^{c/2})$, where $H$ is the Hamiltonian and $c$ is the LD exponent (see Eq.\ \eqref{eq:LD_arc_length}). For the choice $g(x) = 1/x$, we obtain $\rho_U \sim H^{-c/2}$, which emphasizes low-energy regions of the phase space. 

\begin{figure}[htbp]
    \centering
    \includegraphics[width=\textwidth]{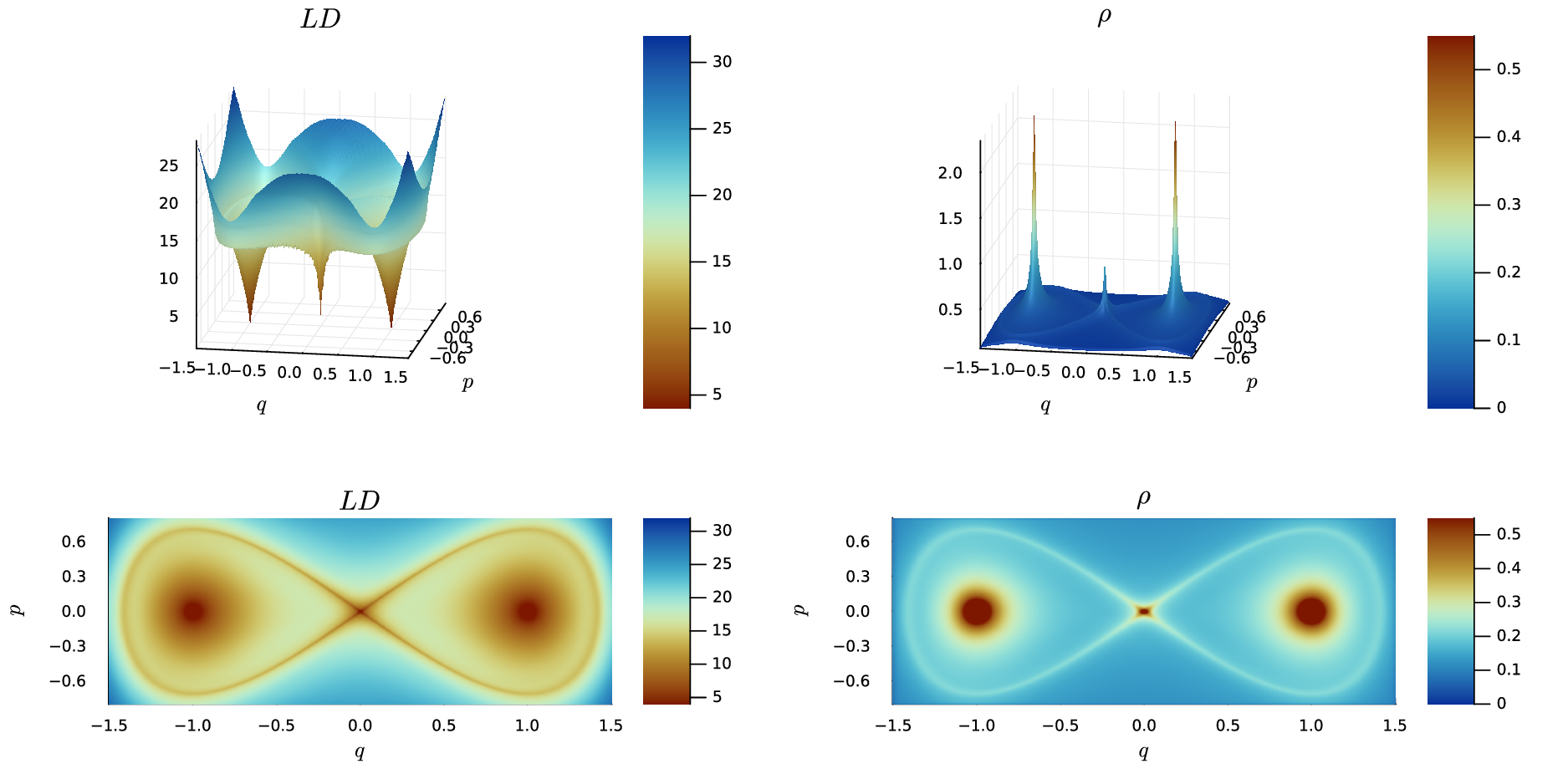}
    \caption[LDs and LD-weighted PDF for the Duffing equation.]
    {LDs and LD-weighted PDF for the Duffing oscillator. 
    Top panels: 3D visualization of LD fields (left) and the corresponding PDF (right). 
    Bottom panels: 2D projections of the same data. 
    Singular features in the LD align with separatrices, while the weighted PDF highlights these regions as dynamically dominant. The LD functional in Eq.~\eqref{eq:LD_arc_length} is used with $c=0.7$ and an integration time of $\tau=4$. The PDF employs the weighting function $g(x) = 1/x$.}
    \label{fig:LD3d}
\end{figure}

Fig.~\ref{fig:LD3d} illustrates the relationship between LD fields and their associated PDFs for the Duffing oscillator (see Eqs.\ \eqref{eq:Duff} below). The left panels show LD values, where homoclinic orbits appear as local minima, marking their role as separatrices in phase space. The right panels display the corresponding PDFs obtained with $g(x) = 1/x$. Both the three-dimensional views (top panels) and the two-dimensional projections (bottom panels) highlight how the LD-weighted density $\rho_U$ emphasizes the geometric structures uncovered by LDs. 

Representative time traces of trajectories initialized in different regions of phase space are shown in Figs.~\ref{fig:time_trace_Duff}–\ref{fig:time_trace_Duff_plus}, illustrating the connection between LD values, the induced PDF weights, and the resulting dynamical behavior. Further discussion of the robustness of the LD-weighted PDFs and their dependence on sampling and integration parameters is provided in Sec.~\ref{sec:sensitivity_analysis}, where the sensitivity of the proposed framework is systematically assessed.

\subsection{Alternative Divergence Measures}

Although the KL divergence \cite{kullback1951information} is our primary tool for comparing LD-weighted probability densities, it is not the only candidate for measuring distributional differences. Several alternative measures are commonly considered:

\begin{itemize}
    \item \textbf{Jensen--Shannon (JS) divergence:} 
    A symmetrized and smoothed version of the KL divergence, defined as 
    \[
    D_{JS}(\rho, \hat{\rho}) = \tfrac{1}{2}D_{KL}(\rho \,\|\, M) + \tfrac{1}{2}D_{KL}(\hat{\rho} \,\|\, M),
    \quad M = \tfrac{1}{2}(\rho + \hat{\rho}).
    \]
    Introduced in \cite{lin2002divergence}, the JS divergence is symmetric and bounded, making it more stable for distributions with disjoint support. However, it is typically less sensitive than KL in emphasizing regions of high probability mass in the reference distribution.

    \item \textbf{Wasserstein distance (Earth mover’s distance):} 
    This optimal transport metric \cite{villani2008optimal, peyre2019computational} measures the minimal ``cost’’ of transporting one distribution into another. It has appealing geometric properties, particularly in capturing differences in the support of distributions. However, its computation requires solving an optimization problem, which becomes expensive in high dimensions and less practical for iterative evaluation in our setting.

\end{itemize}

We selected the KL divergence primarily because it places greater emphasis on regions where the reference distribution $\rho$ is large. In the context of LD-weighted PDFs, this corresponds to dynamically significant regions such as homoclinic orbits. Thus, KL divergence naturally aligns with our goal of assessing how well models reproduce the global phase-space structures most critical to Hamiltonian dynamics. Nonetheless, complementary use of JS divergence or Wasserstein distance may offer valuable robustness checks, which we leave for future investigation.

\section{Data-Driven Models}\label{sec:5:NN_model}

In this section, we discuss the NN architectures employed in this study, with particular emphasis on their ability to preserve the geometric structure of Hamiltonian systems (for further background, see standard references on numerical methods for Hamiltonian systems \cite{Leimkuhler2004,hairer2006GeometricNumerical}). This property is formalized through symplectic geometry, where the canonical symplectic matrix
\begin{equation}
    J = \begin{pmatrix}
        0 & I \\
        -I & 0
    \end{pmatrix},
\end{equation}
plays a foundational role. Here $I$ denotes the $n \times n$ identity matrix. The matrix $J$ satisfies $J^T = -J = J^{-1}$, making it skew-symmetric and orthogonal. A differentiable map $F:U \to U$ is symplectic if its Jacobian preserves the canonical structure through the relation
\begin{equation}
   (\nabla F)^T J (\nabla F) = J,
\end{equation}
which ensures conservation of phase-space volume (Liouville’s theorem) and energy flow. This property is essential for accurately representing physical systems over extended time horizons. NNs that enforce this symplectic condition can therefore preserve critical invariants such as energy, which conventional architectures typically cannot guarantee.

Three architectures---SympNets, H\'enonNets, and GHNN---are explicitly designed to satisfy this symplectic condition. By construction, they admit exact inverses and rigorously conserve the symplectic structure, making them especially well suited for simulating physical systems governed by invariants such as energy and angular momentum.

RC, by contrast, operates outside this geometric framework. Standard RC implementations lack analytical inverses, and therefore require separate models for forward and backward propagation to approximate bidirectional dynamics. Nevertheless, RC’s fixed, randomized reservoir allows efficient adaptation to backward integration, enabling it to approximate Hamiltonian dynamics without explicit symplectic constraints. This flexibility makes RC a compelling alternative, though typically with trade-offs in geometric fidelity compared to symplectic architectures.

The following subsections provide brief overviews of these architectures, with particular focus on their implementation for forward and backward time integration. 

\subsection{SympNets}
SympNets \cite{jin2020SympNetsIntrinsic} are neural architectures explicitly designed to preserve the symplectic structure of Hamiltonian systems. Their core components are \emph{gradient modules}, simple yet powerful building blocks that guarantee symplecticity by construction (see \cite{jin2020SympNetsIntrinsic} for discussion of additional modules). 

For canonical phase-space coordinates $(q,p) \in \mathbb{R}^n \times \mathbb{R}^n$, the upper ($G_u$) and lower ($G_l$) gradient modules are defined as
\begin{equation}
G_u \begin{pmatrix}
    q\\
    p
\end{pmatrix} = 
\begin{bmatrix}
    I & \nabla V \\
    0 & I
\end{bmatrix}
\begin{pmatrix}
    q\\
    p
\end{pmatrix}, 
\qquad 
G_l \begin{pmatrix}
    q\\
    p
\end{pmatrix} = 
\begin{bmatrix}
    I & 0 \\ 
    \nabla V & I
\end{bmatrix}
\begin{pmatrix}
    q\\
    p
\end{pmatrix},
\end{equation}
where $V:\mathbb{R}^n \to \mathbb{R}$ is a learnable potential function acting on either $q$ or $p$. In the original formulation, $V$ is represented by a shallow NN. In our implementation, we parameterize $V$ as
\begin{equation}
  V(q) = a^T \varsigma(Kq+b), 
  \qquad 
  \nabla V(q) = K^T \,\text{diag}(a)\, \sigma(Kq+b), 
  \label{eq:V}
\end{equation}
where $K \in \mathbb{R}^{N_h \times n}$ is a weight matrix, $a \in \mathbb{R}^{N_h}$ a scaling vector, $b \in \mathbb{R}^{N_h}$ a bias, $\sigma$ an activation function, and $\varsigma$ its elementwise antiderivative. The integer $N_h$ denotes the hidden layer width. 

A generalized SympNet (G-SympNet) is constructed by composing such modules:
\begin{equation}
    L = G_{l_k} \circ G_{u_k} \circ \cdots \circ G_{l_1} \circ G_{u_1}.
\end{equation}
SympNets are universal approximators of symplectic maps \cite{jin2020SympNetsIntrinsic}, allowing them to represent arbitrary structure-preserving transformations.

A key advantage of SympNets is their analytical invertibility. Each gradient module admits a simple inverse:
\begin{equation}
    G_u^{-1} = 
    \begin{bmatrix}
        I & -\nabla V \\
        0 & I
    \end{bmatrix}, 
    \qquad 
    G_l^{-1} = 
    \begin{bmatrix}
        I & 0 \\ 
        -\nabla V & I
    \end{bmatrix}.
\end{equation}
Thus, the full network inverse is obtained by reversing the order of module inverses:
\begin{equation}
    L^{-1} = G_{l_1}^{-1} \circ G_{u_1}^{-1} \circ \cdots \circ G_{l_k}^{-1} \circ G_{u_k}^{-1}.
\end{equation}
This exact invertibility enables efficient backward integration of learned dynamics, which is crucial for LD computations requiring bidirectional trajectory integration. By preserving symplectic structure and supporting exact time reversal, SympNets provide a robust and efficient framework for learning Hamiltonian systems. 

The SympNet models are trained by minimizing a mean squared error (MSE) loss of the form
\begin{equation} \label{eq:MSE_loss}
\mathcal{L}_{\mathrm{MSE}} = \frac{1}{N}\sum_{i=1}^{N} \left\| \mathbf{y}_{i+1} - \hat{\mathbf{y}}_{i+1} \right\|^{2},
\end{equation}
where $\mathbf{y}_{i+1}$ denotes the true system state at the next time step and $\hat{\mathbf{y}}_{i+1}$ is the corresponding model prediction.

\subsection{H\'enonNets}
H\'enonNets \cite{burby2020FastNeural} share the same structural philosophy as SympNets, namely the preservation of symplectic structure through invertible modules, but use H\'enon maps as their fundamental building blocks. A H\'enon map is defined as
\begin{equation}
    \hat{h}\begin{pmatrix}
        q\\
        p
    \end{pmatrix} = 
    \begin{bmatrix}
        0 & I \\
        -I & \nabla V
    \end{bmatrix}
    \begin{pmatrix}
        q\\
        p
    \end{pmatrix}
    + \begin{pmatrix}
        0\\
        \eta
    \end{pmatrix},
\end{equation}
where $\eta$ is a constant bias term and $V:\mathbb{R}^n \to \mathbb{R}$ is a learnable scalar potential parameterized by a NN.

A H\'enon layer is constructed by composing four identical H\'enon maps:
\begin{equation}
    L_h = \hat{h} \circ \hat{h} \circ \hat{h} \circ \hat{h},
\end{equation}
and the full H\'enonNet stacks such layers:
\begin{equation}
    L_H = L_{h_k} \circ L_{h_{k-1}} \circ \cdots \circ L_{h_1}.
\end{equation}

The inverse of a single H\'enon map is given by
\begin{equation}
    \hat{h}^{-1}\begin{pmatrix}
        q\\
        p
    \end{pmatrix} = 
    \begin{bmatrix}
        \nabla V & -I \\
        I & 0
    \end{bmatrix}^{-1}
    \begin{pmatrix}
        q\\
        p
    \end{pmatrix}
    + \begin{pmatrix}
        \eta\\
        0
    \end{pmatrix}.
\end{equation}
The full H\'enonNet inverse is obtained by reversing the order of its layer inverses. 

As shown in \cite{horn2025GeneralizedFramework}, H\'enon layers can be interpreted as special cases of the more general G-SympNet framework. This correspondence implies that H\'enonNets inherit the universal approximation guarantees of G-SympNets for symplectic maps, while maintaining the computational efficiencies afforded by their H\'enon map structure. Similarly to SympNet, the H\'enonNet architecture is trained using the MSE loss (Eq.~\eqref{eq:MSE_loss}).

\subsection{Generalized Hamiltonian Neural Network (GHNN)}
The GHNN \cite{horn2025GeneralizedFramework} unifies and extends prior Hamiltonian-preserving architectures, including Hamiltonian NNs (HNNs) \cite{greydanus2019HamiltonianNeural}, Symplectic Recurrent NNs (SRNNs) \cite{Chen2020Symplectic}, SympNets, and H\'enonNets. A key insight is that architectures such as SympNets and H\'enonNets implicitly model separable Hamiltonians through symplectic Euler integration. GHNN generalizes this paradigm by composing arbitrary symplectic maps via modular integration steps:
\begin{equation}
    \text{GHNN}(p,q)
    = \text{SI}_n(\mathcal{H}_n) \circ \text{SI}_{n-1}(\mathcal{H}_{n-1}) \circ \cdots \circ \text{SI}_1(\mathcal{H}_1)(p,q), 
    \qquad 
    \mathcal {H}_i(p,q) = K_i(p) + V_i(q),
\end{equation}
where each $K_i$ and $V_i$ are NNs parameterizing kinetic and potential energy terms, and $\text{SI}_i$ denotes a symplectic integration scheme (e.g., Symplectic Euler, St\"ormer--Verlet). This formulation recovers earlier architectures as special cases:
\begin{itemize}
    \item SympNet arises when $K_i$ and $V_i$ are shallow networks integrated with symplectic Euler.
    \item H\'enonNet corresponds to a constrained GHNN in which pairs of symplectic layers share non-independent parameters.
\end{itemize}
By subsuming these architectures, GHNN provides a flexible and theoretically grounded framework for structure-preserving learning of Hamiltonian dynamics. The GHNN model is likewise trained by minimizing the MSE loss defined in Eq.~\eqref{eq:MSE_loss}.

\subsection{Reservoir Computing (RC)}
RC is a recurrent NN framework in which only the output layer is trainable, while the hidden (reservoir) layer remains fixed after random initialization. First proposed by Jaeger \cite{Jaeger2001}, RC has achieved remarkable success in modeling dissipative systems, including chaotic spatiotemporal prediction \cite{Pathak2018}, multi-scale dynamics \cite{Chattopadhyay2020}, and acoustic resonators \cite{Huhn2021}. Its application to Hamiltonian systems, however, remains relatively unexplored, with limited studies such as \cite{zhangLearningHamiltonianDynamics2021} analyzing Poincar\'e maps of the double pendulum and standard map. 

In general form, RC dynamics are expressed as
\begin{align*}
    x_{k+1} &= f(x_k,u_k), \\
    y_{k+1} &= g(x_{k+1},u_k),
\end{align*}
where $u_k \in \mathbb{R}^{N_u}$ is the input at time step $k$, $x_k \in \mathbb{R}^{N_h}$ is the hidden reservoir state with $N_h \gg N_u$, $f$ governs the fixed reservoir dynamics, and $g$ is the trainable output layer mapping hidden states and inputs to the output $y_{k+1}$.

The universal approximation capability of RC has been established in several theoretical works \cite{grigoryevaEchoStateNetworks2018,gonon2023ApproximationBounds,hartEmbeddingApproximationTheorems2020,hartEchoStateNetworks2021}, guarantee that RC can approximate wide class of dynamical systems. The Echo State Property (ESP) \cite{Jaeger2001,yildizRevisitingEchoState2012} guarantees that the reservoir asymptotically ``forgets'' initial conditions: after sufficiently long input sequences, $x_k$ becomes independent of initialization. In practice, an initial transient ($N_w$ warm-up steps) is discarded during training and prediction to eliminate initialization artifacts.

In this study, the reservoir dynamics follow the leaky integration form:
\begin{align}
    f(x_k,u_k) &= (1-\alpha) x_k + \alpha \tanh(W_x x_k + W_u u_k), \\
    g(x_{k+1},u_k) &= W x_{k+1}, \label{eq:output_layer}
\end{align}
where $0<\alpha \leq 1$ is the leaky integration rate, $W_x \in \mathbb{R}^{N_h\times N_h}$ is the adjacency matrix (typically sparse) defining reservoir connections, and $W_u \in \mathbb{R}^{N_h\times N_u}$ is the input weight matrix. The entries of $W_x$ are randomly generated and rescaled to have maximum spectral radius $\rho$, while the entries of $W_u$ are uniformly distributed in $[-\sigma_B,\sigma_B]$. Nonlinearity is introduced elementwise via the hyperbolic tangent function $\tanh(\cdot)$.

Training reduces to fitting the output weights $W$ by minimizing
\begin{equation}
    L(g) = \sum_{k\in\mathcal{K}} \|y_k - \hat{y}_k \|^2 + \beta_L \|g \|,
\end{equation}
where $\mathcal{K}$ is the training index set, $\hat{y}_k$ are the target outputs, and $\beta_L$ is a regularization parameter. Apart from the regularization parameter, this loss function differs from the mean squared error in Eq.~\eqref{eq:MSE_loss}, used in the symplectic NN models, only by a normalization factor. Unlike general RNNs, RC training is equivalent to a regularized least-squares problem, solvable either by pseudoinverse or gradient-based optimization. This computational simplicity is a major advantage, enabling efficient training even on large datasets. Once trained, RC operates autonomously by recursively feeding predictions $y_k$ back as inputs $u_k$. 

The performance of RC is highly sensitive to hyperparameters. In our experiments, we fix the hidden layer size $N_h$ and sparsity $\mu$ of $W_x$, while optimizing the hyperparameters $\alpha, \beta_L, \sigma_B,$ and $\rho$ using the Covariance Matrix Adaptation Evolution Strategy (CMA-ES) \cite{hansen2001CompletelyDerandomized}. At each CMA-ES iteration, a new RC model is trained and evaluated on a validation set using multi-step prediction error. The best-performing model is selected, ensuring generalization to unseen trajectories. Optimization terminates after a fixed computational budget is reached.

For LD computations, a backward operator is required. Because RC lacks an analytical inverse, we construct a backward model by time-reversing the training data. Specifically, after training the forward RC via CMA-ES, we keep the hidden layer parameters ($\alpha, W_x, W_u$) and $\beta_L$ fixed, and retrain the output layer on reversed trajectories. This approach exploits Hamiltonian time-reversibility and yields an efficient backward approximation essential for bidirectional LD integration.

LD computation then proceeds via autonomous prediction. Each trajectory is initialized by a warm-up phase, integrating backward for $N_w$ steps to reach a starting state $\tilde{u}_0$ in phase space. Both forward and backward predictions are then used to compute LDs, ensuring accurate reconstruction of global structures.

\section{Numerical Experiment: Duffing Equation}\label{sec:5:Duffing}
The Duffing oscillator provides a paradigmatic benchmark for evaluating NN performance in Hamiltonian dynamics. This system exhibits homoclinic orbits and complex fixed-point topology, making it an ideal test case for structure-preserving learning. We begin by presenting its governing equations and characteristic phase-space structures. After detailing our dataset generation methodology and training protocols, we describe the NN architectures and hyperparameters. The analysis proceeds in three stages: (i) short-term trajectory forecasting across representative regions of phase space, (ii) LD-based visualization and quantification of phase-space reconstruction, and (iii) systematic investigation of how training dataset size influences reconstruction fidelity. The LD framework thus reveals structural deficiencies that conventional trajectory-based error metrics fail to capture.

\subsection{Governing Equation and Training Data}
The Duffing equation is a canonical nonlinear oscillator characterized by cubic nonlinearity, making it an ideal textbook example for studying Hamiltonian dynamics. Its mathematical simplicity, combined with rich dynamical behavior, provides an excellent benchmark for evaluating NN architectures. Furthermore, several soliton equations can be reduced to the Duffing form under appropriate transformations \cite{kovacic2011duffing}. 

The integrable Duffing equation considered in this study is governed by the separable Hamiltonian
\begin{equation}\label{eq:Ham_Duffing}
    \mathcal{H}_{D}(q,p) = \frac{1}{2}p^2 + V_D(q), 
    \qquad V_D(q) = -\frac{1}{2}q^2 + \frac{1}{4} q^4,
\end{equation}
where the kinetic energy term $\tfrac{1}{2}p^2$ and the potential energy $V_D(q)$ are explicitly decoupled. The corresponding Hamiltonian equations of motion are
\begin{subequations}
\begin{align}
    \dot{q} &= p, \label{eq:Duffq}\\
    \dot{p} &= -q + q^3. \label{eq:Duffp}
\end{align}
\label{eq:Duff}
\end{subequations}This separable structure makes the Duffing equation particularly well-suited for benchmarking GHNNs, H\'enonNets, and SympNets, since these architectures are designed to exploit separable Hamiltonians in enforcing energy conservation and symplecticity.

The phase space of this system contains three fixed points: two centers at $(q,p) = (\pm 1,0)$ and one saddle point at $(q,p) = (0,0)$. Most relevant to our analysis is the homoclinic orbit associated with the saddle point. The energy level $\mathcal{H}_D = 0$ defines a figure-eight separatrix that connects the saddle to itself and encloses the two centers. This homoclinic orbit admits the analytical representation
\begin{equation}
    p = \pm q\sqrt{1-\tfrac{1}{2}q^2},
\end{equation}
which describes trajectories that asymptotically approach the saddle point both forward and backward in time.

\subsection{Dataset Generation}
Training data are generated by numerically integrating the Duffing system \eqref{eq:Duffq}--\eqref{eq:Duffp} with initial conditions $u_0 = (p(0),q(0))=(0,Q_0)$, where $Q_0$ is sampled uniformly from $(-3,3)$. This sampling ensures coverage of trajectories both inside and outside the homoclinic orbit. Each trajectory is integrated for 100 time units with a fixed output step size $\Delta t = 0.1$, yielding 1000 time points per trajectory. We collect 200 trajectories in total and split them into training (80\%), validation (10\%), and testing (10\%) sets. Throughout this section, we refer to this configuration as the \emph{200-trajectory case}.

Numerical integration is performed using the \texttt{DifferentialEquations.jl} package, employing its adaptive autoswitching algorithm. Specifically:
\begin{itemize}
    \item The Tsitouras 5/4 Runge--Kutta method is used for non-stiff regions,
    \item A fourth-order Rosenbrock method with a stiff-aware third-order interpolant is used for stiff regimes.
\end{itemize}

\subsection{Training and Quality Metrics}\label{subsec:Duffing_NN_LD_calculation}
Ensuring a fair comparison between different NN architectures is nontrivial. Execution time, while intuitive, is not reliable because it depends heavily on implementation details such as hardware, compiler optimizations, and memory access patterns. This issue is particularly relevant here, as the models were implemented in different languages: Python for SympNet, H\'enonNet, and GHNN (following \cite{horn2025AELITTENGHNN}), and Julia for RC (following \cite{martinuzzi2022reservoircomputing}).

To avoid these biases, we compare models in terms of computational complexity, specifically the number of floating-point operations (FLOPs) required during inference. This metric reflects the intrinsic efficiency of the architectures, independent of implementation details. The derivation of FLOP counts is provided in Appendix~\ref{App:CompCost}.

For consistency, SympNets and H\'enonNets are implemented with a single-layer MLP potential, while GHNNs employ symplectic Euler integration with separate two-layer MLPs parameterizing kinetic and potential terms. These symplectic architectures are trained with the Adam optimizer (learning rate $10^{-3}$, $\beta_1 = 0.9$, $\beta_2 = 0.999$) for 250 epochs. RC is implemented with 400 hidden neurons and sparsity $\mu=0.006$, resulting in an average connectivity of 2.4 per neuron. The detailed configurations and FLOP counts are reported in Table~\ref{tab:nn_config_duffing}.

\begin{table}[htbp]
    \centering
    \small
    \begin{tabular}{lccccc}
        \hline
        NN type & \makecell{Learned\\ Hamiltonians} & Layers & \makecell{Neurons \\ per Layer} & \makecell{Trainable \\ parameters} & \makecell{\#FLOPs\\ (Inference)}\\
        \hline
        SympNet & 10 & 1 & 50 & 3000 & 6000 \\
        H\'enonNet & 10 & 1 & 50 & 755 & 6020 \\
        GHNN & 3 & 2 & 15 & 1710 & 6120 \\
        RC & -- & -- & 400 & 800 & 6318 \\
        \hline
    \end{tabular}
    \caption{NN configurations and FLOP counts for the Duffing equation.}
    \label{tab:nn_config_duffing}
\end{table}

\begin{figure}
    \centering
    \includegraphics[width=0.9\linewidth]{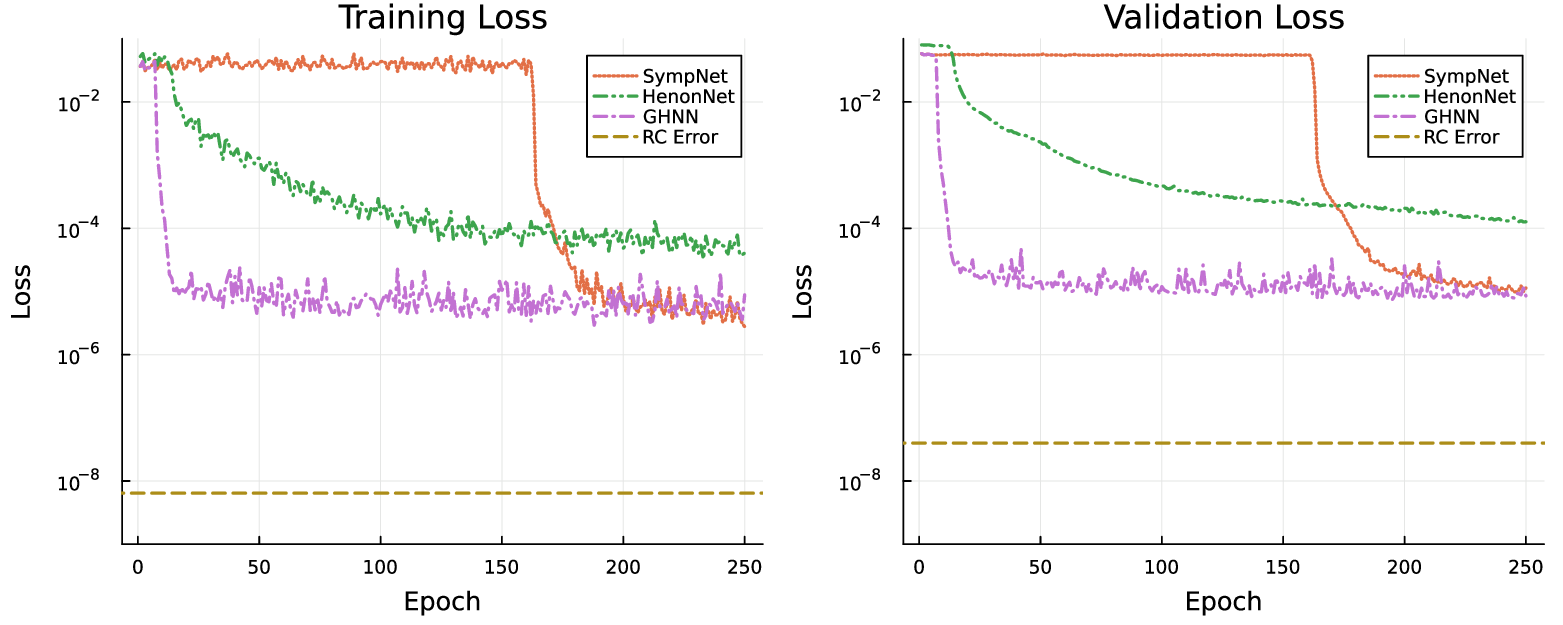}
    \caption{Training (left) and validation (right) loss for the NN models in the Duffing case. 
SympNet (red dotted), H\'enonNet (green dash–dotted), GHNN (purple dash–dotted), and Reservoir Computing (RC, orange dashed). 
Note that RC training does not involve iterative optimization; the output weights are obtained by solving a regularized normal equation.}
    \label{fig:loss_compare_duffing}
\end{figure}

The training and validation loss for the Duffing case is shown in Fig.~\ref{fig:loss_compare_duffing}. Among the symplectic-type architectures, GHNN exhibits the fastest convergence in this experiment, reaching a normalized mean squared error (NMSE) of approximately $10^{-5}$ within only a few epochs. SympNet displays a distinct transition around epoch 150, after which the loss rapidly decreases and convergence is achieved. In contrast, H\'enonNet shows a more gradual reduction of the loss and does not reach the same error level within the maximum training horizon of 250 epochs.

For comparison, we also report the NMSE obtained with Reservoir Computing (RC), which is on the order of $10^{-8}$. It should be noted, however, that RC is trained in teacher-forcing mode and the output weights are obtained by solving a regularized normal equation rather than through iterative gradient-based optimization, meaning that RC training does not involve training epochs. 

To evaluate trajectory accuracy, we introduce a normalized error metric:
\begin{equation}
    \hat{e}(u_0) = \int_{-\tau}^{\tau} \big(|q(t)-\hat{q}(t)|^c + |p(t)-\hat{p}(t)|^c\big)\,dt, \quad e(u_0) = 
    \frac{\hat{e}(u_0)}{LD(u_0)}, 
    \qquad c=0.7,
    \label{eq:error_Duffing}
\end{equation}
where $(q(t),p(t))$ and $(\hat{q}(t),\hat{p}(t))$ denote the ground-truth and model-predicted trajectories, respectively, and $\tau$ is the integration horizon. The denominator normalizes by the LD
\begin{equation}
    LD(u_0) = \int_{-\tau}^{\tau} \big(|\dot{q}|^c+ |\dot{p}|^c\big)\,dt,
    \label{eq:LD_Duffing}
\end{equation}
ensuring comparability of errors across different regions of phase space with varying dynamical activity. Time derivatives are approximated using forward differences. The reference distribution $\rho_{\text{ref}}$ in Fig.~\ref{fig:LD3d} was computed from direct numerical integration of Eqs.~\eqref{eq:Duffq}--\eqref{eq:Duffp}.

\subsection{Short-time forecasting}
\begin{figure}
    \centering
    \includegraphics[width=\linewidth]{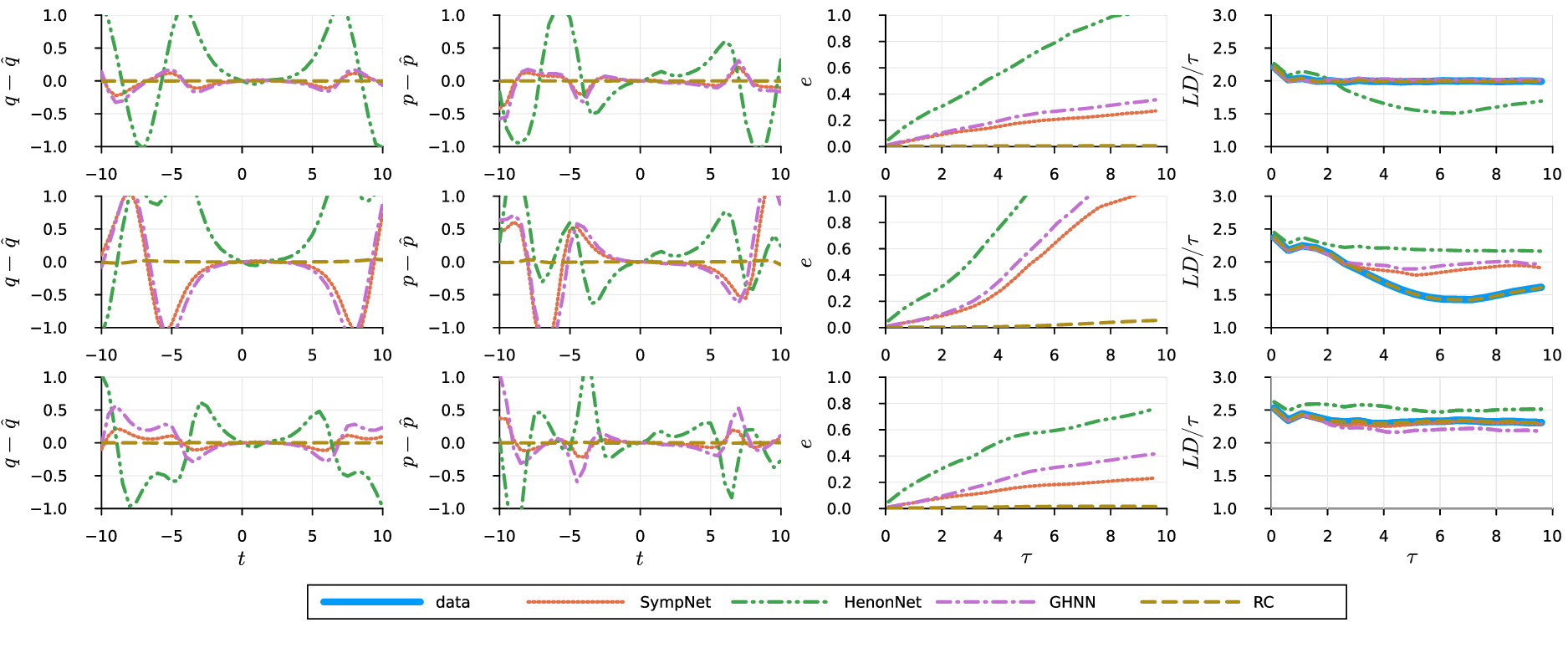}
    \caption[Forward and backward predictions for Duffing equation.]{Forward and backward predictions of the Duffing equation starting from $t_0=0$. Top row: $(q(0), p(0))=(-0.71, -0.51)$ (inside homoclinic orbit); Middle row: $(q(0), p(0))=(-0.71, -0.61)$ (near homoclinic orbit); Bottom row: $(q(0), p(0))=(-0.1, -0.71)$ (outside homoclinic orbit). From left to right, the columns show: the pointwise error in position $q(t)-\hat{q}(t)$, the pointwise error in momentum $p(t)-\hat{p}(t)$, the normalized prediction error (Eq.~\eqref{eq:error_Duffing}) as function of the integration time $\tau$, and normalized LD (Eq.~\eqref{eq:LD_Duffing}) as function of the integration time $\tau$. Predictions are shown over the interval $t\in[-10,10]$. The reference solution (blue solid) is compared with SympNet (red dotted), H\'enonNet (green dash-dotted), GHNN (purple dash-dotted), and RC (orange dashed). All NNs were trained on the dataset corresponding to the 200-trajectory case.}
    \label{fig:time_trace_Duff}
\end{figure}

Figure \ref{fig:time_trace_Duff} presents predictions for three representative initial conditions—inside, near, and outside the homoclinic orbit—highlighting clear differences in NN performance across these dynamical regimes. RC consistently achieves the highest accuracy, reproducing the reference solution with excellent agreement even in the sensitive near-homoclinic region. SympNet and GHNN perform well for trajectories inside and outside the homoclinic orbit but show substantial deviations near the boundary, where the dynamics are most sensitive. H\'enonNet exhibits the largest discrepancies across all regimes, with particularly strong errors in the vicinity of the homoclinic orbit.

The character of these deviations is informative. H\'enonNet predictions preserve the amplitude range of the reference solution but fail to reproduce the correct temporal structure, oscillating at incorrect frequencies and phases. When initialized near the homoclinic orbit, H\'enonNet erroneously generates dynamics characteristic of trajectories outside the orbit, thereby missing the properties of the homoclinic structure. In contrast, near-homoclinic predictions from SympNet and GHNN resemble trajectories inside the orbit, indicating that these NNs systematically misplace the homoclinic boundary. Only RC captures the qualitative features of near-homoclinic trajectories, although minor phase discrepancies emerge at longer prediction horizons. These qualitative differences are not fully reflected by the normalized prediction error (third column), since pointwise error measures can obscure structural deviations in trajectory geometry.

\begin{figure}
    \centering
    \includegraphics[width=\linewidth]{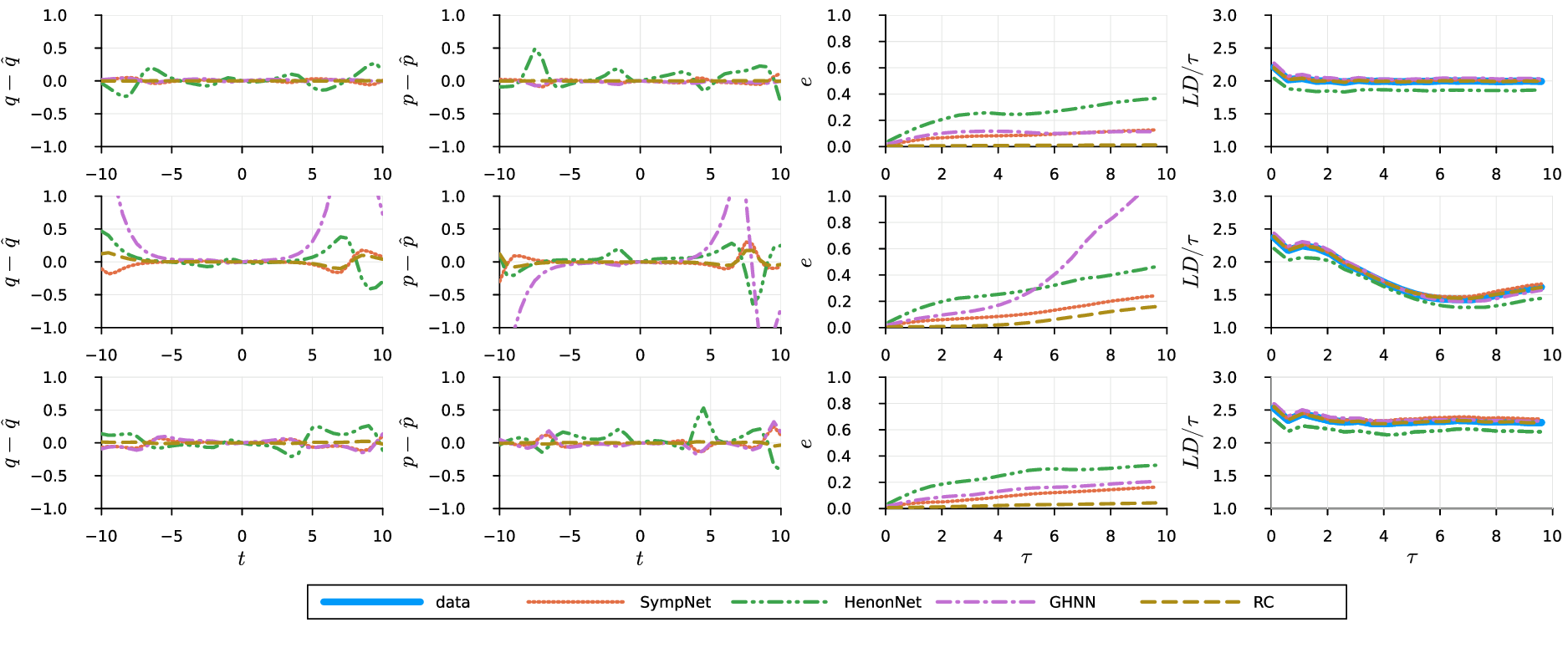}
    \caption[Similar to Fig. \ref{fig:time_trace_Duff} but with different initial conditions.]{Same as Fig.~\ref{fig:time_trace_Duff}, but with $p(0)>0$. Top row: $(q(0), p(0))=(-0.71, 0.51)$ (inside homoclinic orbit); Middle row: $(q(0), p(0))=(-0.71, 0.61)$ (near homoclinic orbit); Bottom row: $(q(0), p(0))=(-0.1, -0.71)$ (outside homoclinic orbit).}
    \label{fig:time_trace_Duff_plus}
\end{figure}

The second set of experiments (Fig.~\ref{fig:time_trace_Duff_plus}) examines symmetric initial conditions with identical positions but opposite momentum ($p(0)>0$ instead of $p(0)<0$). In these cases, all NNs perform better, with even H\'enonNet producing qualitatively accurate predictions. The exception is GHNN near the homoclinic orbit (middle panels), which remains accurate initially but develops substantial deviations beyond the prediction horizon $|t|>5$. This asymmetric behavior across phase space indicates non-uniform learning capability among the architectures.

\begin{figure}
    \centering
    \includegraphics[width=0.49\linewidth]{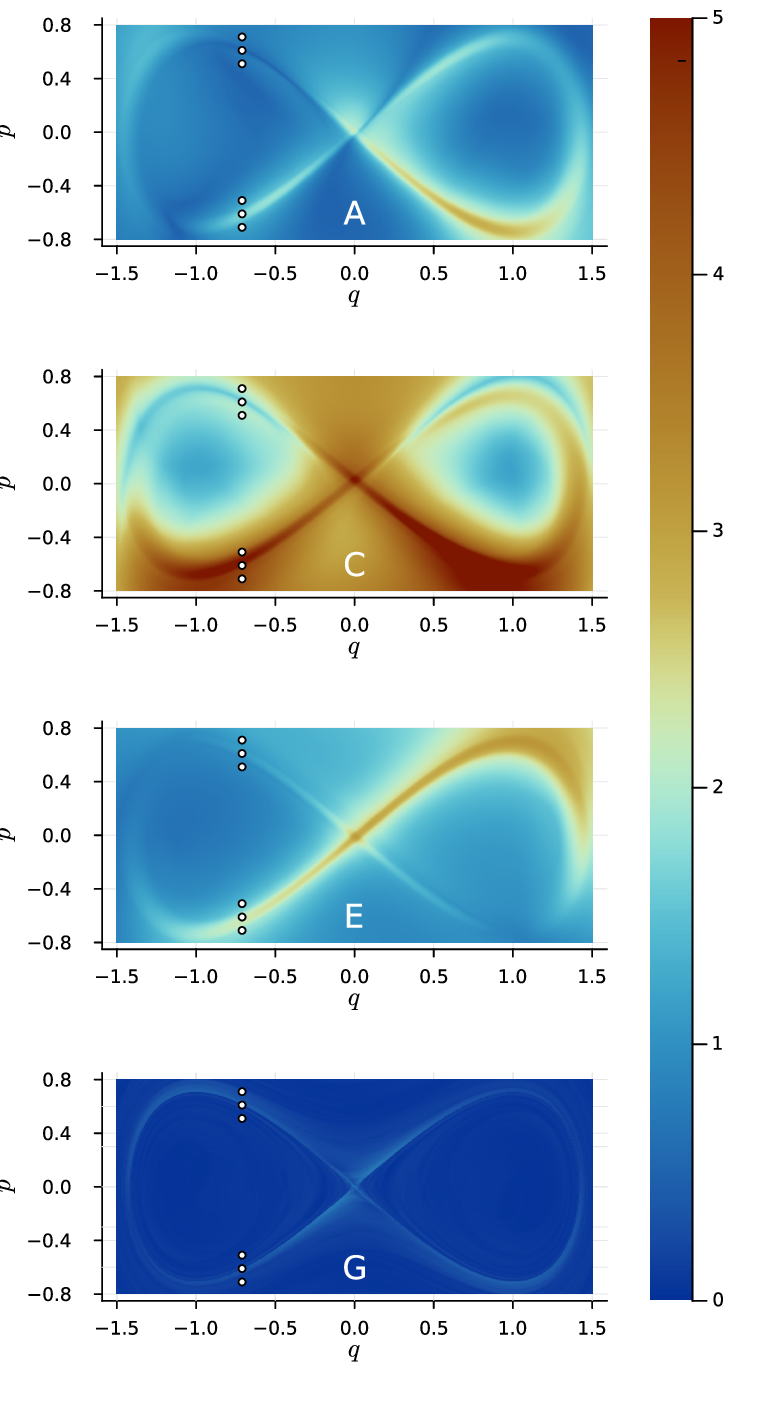}
    \includegraphics[width=0.49\linewidth]{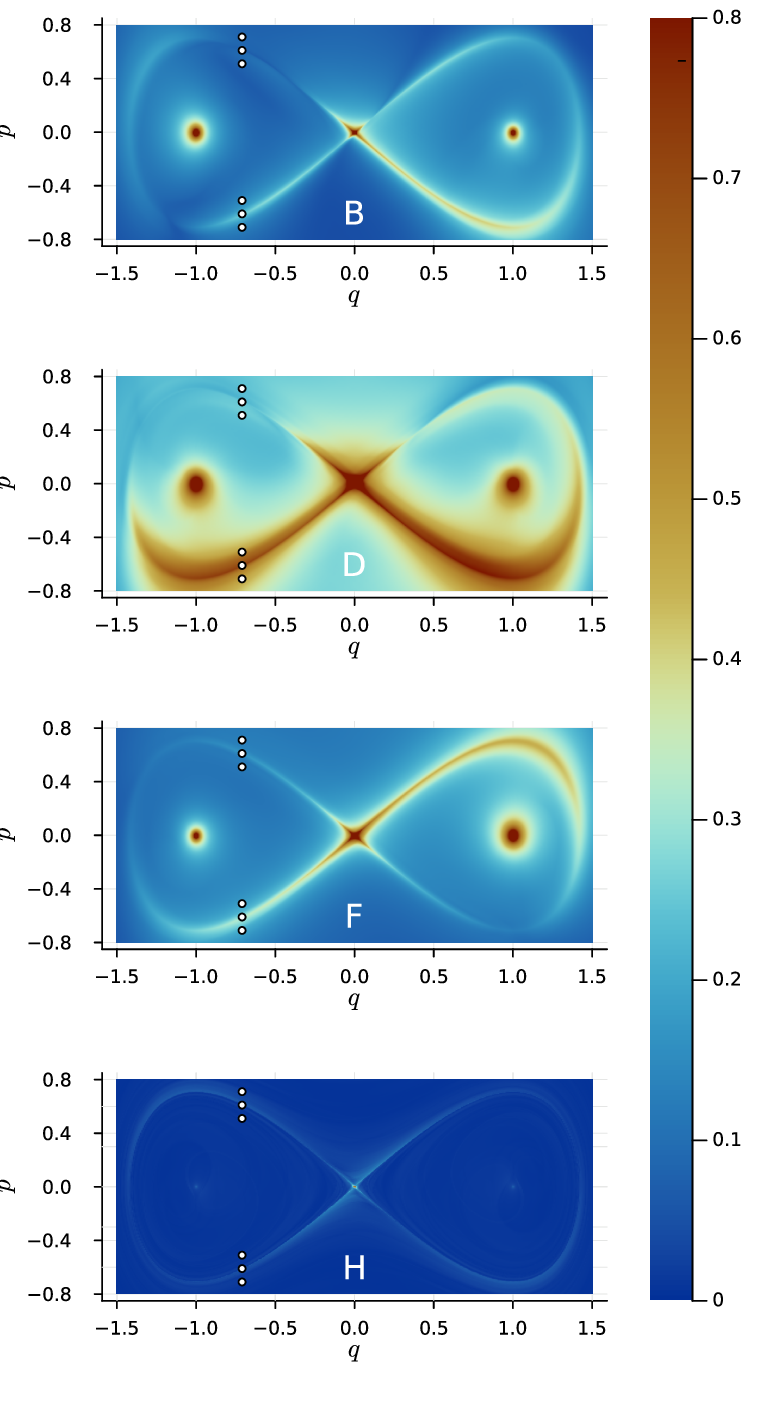}
    \caption{The prediction error $\hat{e}$ (left) and normalized prediction error $e$ (right) of Duffing trajectories across initial conditions : (A-B) SympNet, (C-D) H\'enonNet, (E-F) GHNN, (G-H) RC. The white dots indicate the initial conditions of the trajectories shown in Figs.~\ref{fig:time_trace_Duff} and \ref{fig:time_trace_Duff_plus}.}
    \label{fig:LD_err_Duffing}
\end{figure}

To evaluate performance systematically across the phase space, we compute the normalized prediction error (Eq.~\ref{eq:error_Duffing}) for integration time $\tau=4$ on a uniform grid with $q(0)\in[-1.5,1.5]$ and $p(0)\in[-0.8,0.8]$ (400 points in each direction). Fig.~\ref{fig:LD_err_Duffing} shows the normalized error distributions (see also the corresponding non-normalized error in Fig.\ref{fig:LD_Duffing}), revealing distinct patterns for each architecture. RC attains consistently small errors, demonstrating robust generalization across dynamical regimes. SympNet and GHNN remain accurate away from critical structures but develop substantial errors near the homoclinic orbit and around the fixed points $(q,p)=(\pm 1,0)$. H\'enonNet produces the largest errors overall, particularly in the lower half-plane of momentum space and along the homoclinic orbit. 

\begin{figure}
    \centering
    \includegraphics[width=0.5\linewidth]{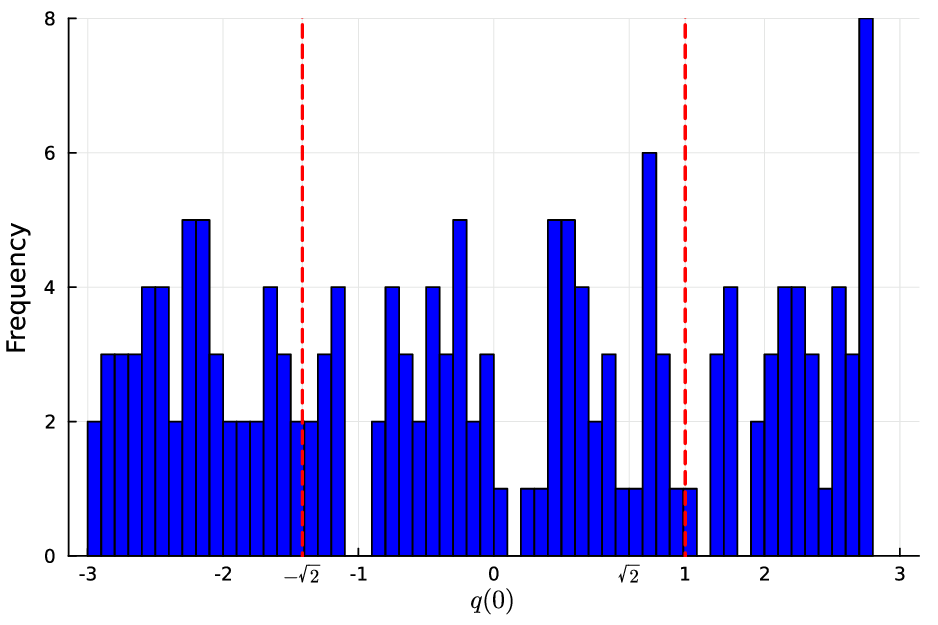}
    \caption{Distribution of initial conditions $q(0)$ in the training dataset for the 200-trajectory case. The red vertical lines indicate $q(0)=\pm\sqrt{2}$, corresponding to the intersections of the homoclinic orbit with the line $p=0$}
    \label{fig:duffing_ic_distribution}
\end{figure}

It is interesting to examine the source of the asymmetry observed in the learned model error of Fig.~\ref{fig:LD_err_Duffing}. We attribute this effect to two main factors. First, although the training data were generated uniformly over $q(0)\in[-1.5,1.5]$, the realized dataset exhibits slight asymmetry. In particular, a greater number of initial conditions lie near $q(0)\approx 0^{-}$ and $-\sqrt{2}$ than near $0^{+}, \sqrt{2}$ (see Fig.~\ref{fig:duffing_ic_distribution}). These locations correspond to the intersections of $p(0)=0$ with the homoclinic orbit, so even a modest sampling imbalance can influence how accurately the associated invariant structures are learned. Second, because the symplectic NN models are trained only in the forward time direction, structures associated with backward-time dynamics—such as the stable manifolds—are less accurately captured. The combination of data imbalance and one-directional training likely contributes to the observed asymmetric error distribution in the learned symplectic models.

These error distributions highlight each NN's ability to capture the global phase-space topology. As shown in the next section, while all architectures identify the existence of the homoclinic orbit, their accuracy in resolving its geometry and the surrounding dynamical structures differs substantially—differences that correspond directly to the error patterns observed here.

\subsection{LDs calculation}
This section demonstrates how LDs reveal critical dynamical features of the Duffing system and provide insight into the performance disparities among NNs.

Figs.~\ref{fig:time_trace_Duff}--\ref{fig:time_trace_Duff_plus} illustrate LD calculations for representative initial conditions as functions of integration time $\tau$. For trajectories inside and outside the homoclinic orbit, the normalized LD values ($LD/\tau$) rapidly converge to stable asymptotic values, consistent with regular quasi-periodic motion. In contrast, trajectories initialized near the homoclinic orbit exhibit pronounced fluctuations in normalized LD values, reflecting the slow nonlinear dynamics characteristic of this boundary.

To evaluate the global phase-space structure, we compute LDs for each trained NN on a uniform grid with $q(0)\in [-1.5,1.5]$ and $p(0)\in [-0.8,0.8]$ (400 points per direction). The integration time is $\tau=4$. From these LD values, we construct PDFs using the weighting function $g(x)=1/x$, which emphasizes dynamically significant regions.

\begin{figure}
    \centering
    \includegraphics[width=0.49\linewidth]{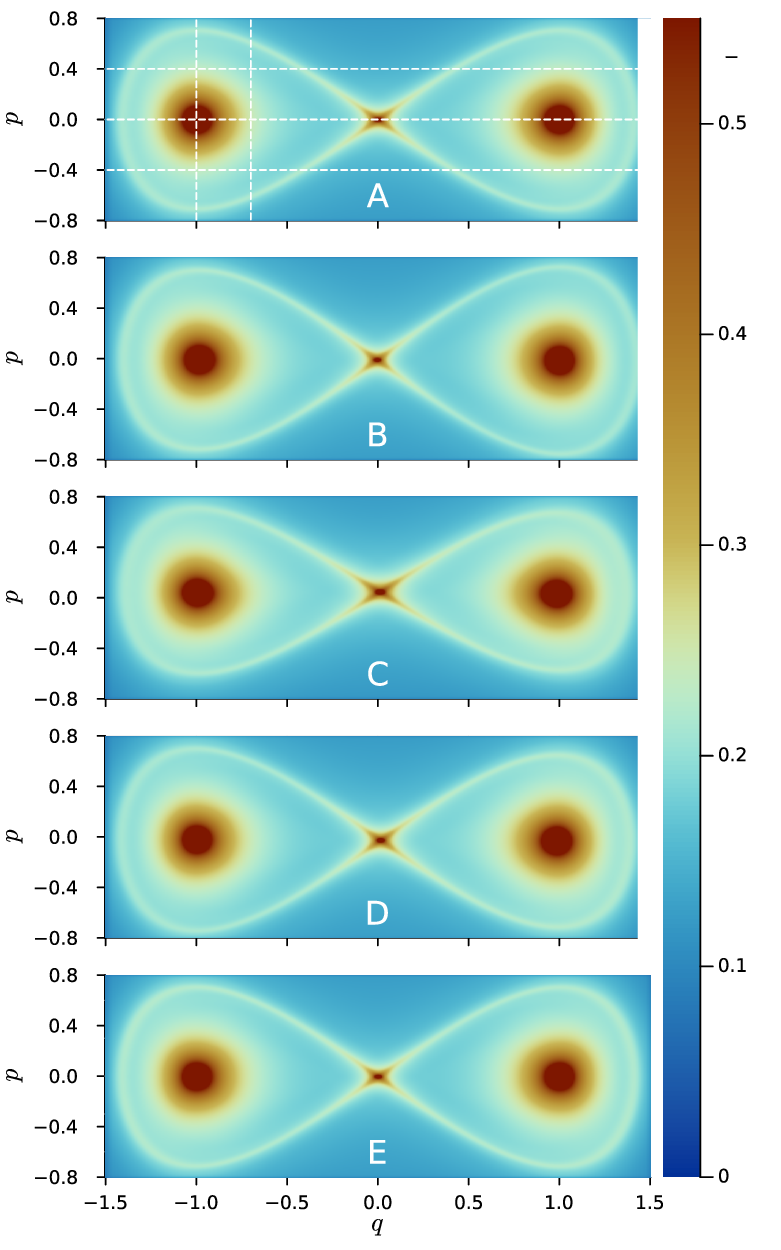}
    \includegraphics[width=0.49\linewidth]{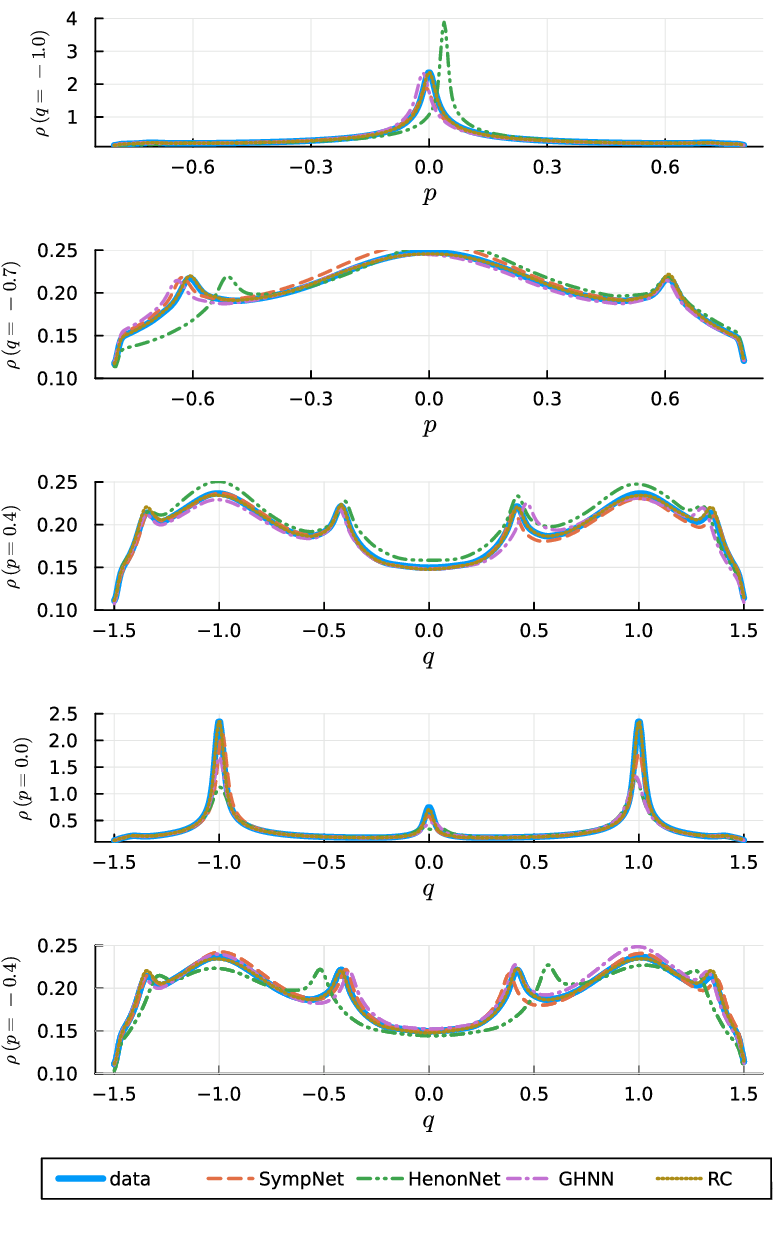}
    \caption[LD-weighted PDF for Duffing equation.]{LD-weighted PDFs for the Duffing equation. The left panels: (A) Numerical integration, (B) SympNet, (C) H\'enonNet, (D) GHNN, (E) RC. All NNs were trained on the dataset corresponding to the 200-trajectory case. High-density regions correspond to slow dynamics, revealing the fixed points near $(q,p)=(\pm1,0)$.  The white lines indicate the phase-space cross sections shown in the right panels. Right panels: one-dimensional slices of the LD-weighted PDF across all methods at selected phase-space locations. 
From top to bottom: $q=-1.0$, $q=-0.7$, $p=-0.4$, $p=0.0$, and $p=0.4$.}
    \label{fig:LD_PDF_Duffing}
\end{figure}

Fig.~\ref{fig:LD_PDF_Duffing} shows the resulting LD-weighted PDFs. All NNs identify the Duffing system’s fundamental topological features: two center fixed points at $(q,p)\approx (\pm1,0)$, a saddle point at $(q,p)\approx (0,0)$, and the connecting homoclinic orbit. Numerical LDs (top panel) are shown for reference. 

\begin{figure}
    \centering
    \includegraphics[width=0.8\linewidth]{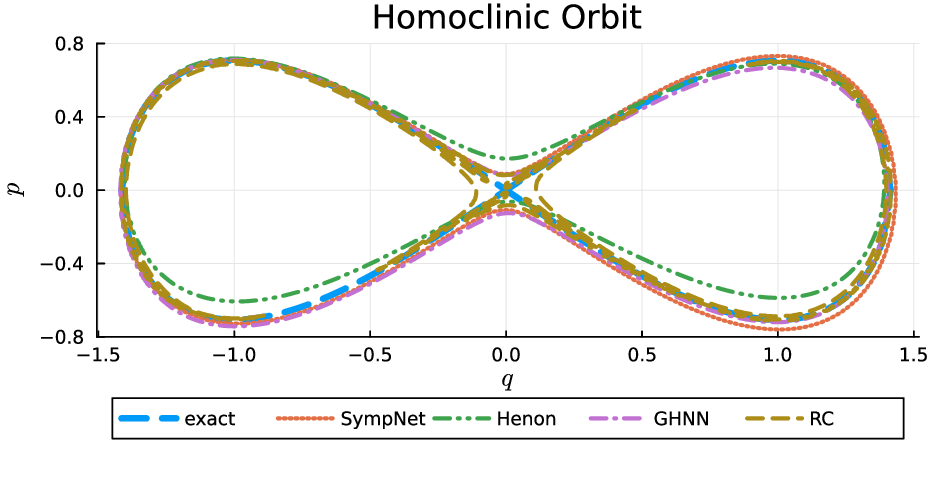}
    \caption[Comparison of homoclinic orbits predicted by NN models against the exact Duffing solution.]{Comparison of homoclinic orbits predicted by NNs against the exact Duffing solution (Eq.~\eqref{eq:Ham_Duffing}, blue dashed). SympNet (red dotted), H\'enonNet (green dash-dotted), GHNN (purple dash-dotted), and RC (brown dashed). LDs reveal deviations in the homoclinic structure: H\'enonNet’s predicted orbit diverges significantly from the reference solution, while the other NNs align more closely. All models were trained on the dataset corresponding to the 200-trajectory case.}
    \label{fig:homoclinic_orbit}
\end{figure}

Although visual inspection suggests qualitative agreement across models, closer examination reveals important geometric differences in the homoclinic orbit (Fig.~\ref{fig:homoclinic_orbit}). In particular, while the symplectic NN models reproduce the existence of the homoclinic orbit, its precise location is slightly shifted relative to the ground truth. Because the LD-based PDF reflects global transport barriers and invariant manifold structure rather than pointwise trajectory alignment, modest displacements of the separatrix may leave the overall PDF largely unchanged. However, even small geometric shifts of the homoclinic orbit can lead to significant local trajectory discrepancies, especially near this sensitive boundary where the dynamics strongly amplify perturbations. These localized deviations are precisely what is observed in the prediction error shown in Fig.~\ref{fig:LD_err_Duffing}.

A detailed comparison of the homoclinic orbits is presented in Fig.~\ref{fig:homoclinic_orbit}, where the orbits are extracted using a level-set method based on the marching squares algorithm \cite{lorensen1987MarchingCubes}, implemented via the Contour.jl package \cite{contourjl2024}. RC (brown dashed) closely reproduces the orbit’s geometry, SympNet (red dotted) and GHNN (purple dash-dotted) exhibit modest curvature deviations, while H\'enonNet (green dash-dotted) shows severe distortions, particularly in the lower half-plane of momentum space.

Quantitative comparison via the KL divergence (Table~\ref{tab:kl_duffing}) further supports these observations. RC achieves the smallest divergence value, indicating the most accurate representation of the global phase-space topology. GHNN and SympNet attain slightly larger but still relatively low divergence values, reflecting modest geometric deviations. In contrast, H\'enonNet exhibits the largest divergence, confirming its inability to resolve the Duffing system’s global structure.

The LD analysis clarifies the trajectory prediction results. RC’s accurate resolution of the homoclinic orbit explains its exceptional short-time forecasting performance, particularly near this sensitive boundary where dynamics are most susceptible to perturbations. The modest geometric discrepancies in SympNet’s and GHNN’s homoclinic orbits correspond to their intermediate prediction errors, while H\'enonNet’s severe distortions account for its persistent failures across all dynamical regimes. This correspondence between phase-space geometry and trajectory accuracy demonstrates how LD analysis connects local prediction behavior with global dynamical structure, providing a powerful diagnostic tool for assessing NNs trained on Hamiltonian systems.

\subsection{Effect of training data}
This section examines the relationship between training dataset size and NN performance, showing how LDs provide unique insights into phase-space learning dynamics. Each architecture was trained on datasets containing $N = \{10, 20, 50, 100, 200, 300, 400, 500\}$ trajectories.  The datasets are not nested; initially, 500 trajectories were generated, and smaller datasets were created by randomly selecting trajectories from the larger dataset. For each trained model, we calculated LD-weighted PDFs using the same procedure as before and quantified deviations from the reference distribution using the KL divergence (Table~\ref{tab:kl_duffing}).

\begin{table}[htbp!]
\centering
   \scriptsize
\begin{tabular}{lcccccccc}
    \hline
\multirow{2}{*}{NN models} & \multicolumn{8}{c}{Number of training trajectories $N$} \\
\cline{2-9}
 & 10 & 20 & 50 & 100 & 200 & 300 & 400 & 500 \\
\hline
SympNet   & 2.17e-01 & 2.12e-01 & 2.19e-01 & 2.13e-01 & 1.32e-03 & 3.24e-04 & 1.01e-03 & 6.87e-04 \\
H\'enonNet & 1.76e-01 & 1.80e-01 & 1.84e-02 & 1.07e-02 & 8.03e-03 & 4.64e-03 & 4.16e-04 & 8.04e-03 \\
GHNN      & 2.17e-01 & \textbf{7.11e-03} & \textbf{6.30e-04} & \textbf{1.02e-03} & 2.75e-03 & 2.71e-04 & \textbf{1.74e-04} & \textbf{1.28e-04} \\
RC        & \textbf{1.07e-01} & 6.16e-01 & 3.59e-01 & 1.02e-01 & \textbf{6.20e-05} & \textbf{1.10e-04} & 8.03e-03 & 8.40e-04 \\
\hline
\end{tabular}
\caption[KL divergence between model-predicted and reference LD-weighted PDFs.]{KL divergence between model-predicted and reference LD-weighted PDFs for the Duffing equation, evaluated across training dataset sizes $N=\{10,20,50,100,200,300,400,500\}$. Lower values indicate better reconstruction of the global phase-space structure.}
\label{tab:kl_duffing}
\end{table}

The results reveal architecture-specific sensitivity to training dataset size. GHNN demonstrates strong data efficiency, outperforming other models both for small datasets ($N=20$--50) and in the large-data regime ($N>300$). This effectiveness likely stems from its explicit symplectic structure combined with deeper parametrization of the Hamiltonian. Interestingly, RC achieves its best performance for intermediate dataset sizes ($N=200$--300$)$, where it slightly outperforms GHNN despite lacking built-in physical constraints.

RC’s diminished relative performance for larger datasets ($N>300$) is counterintuitive. Two factors may contribute: (i) the simpler architecture of RC may impose a ceiling on representational capacity, preventing it from fully exploiting additional training data; (ii) the fixed stopping time used in hyperparameter optimization may disadvantage RC for large datasets, since each evaluation requires more computation and leaves fewer opportunities for parameter exploration. These results highlight the complex interplay between architecture, physical constraints, and training methodology in learning Hamiltonian systems.

\begin{figure}
    \centering
    \includegraphics[width=\linewidth]{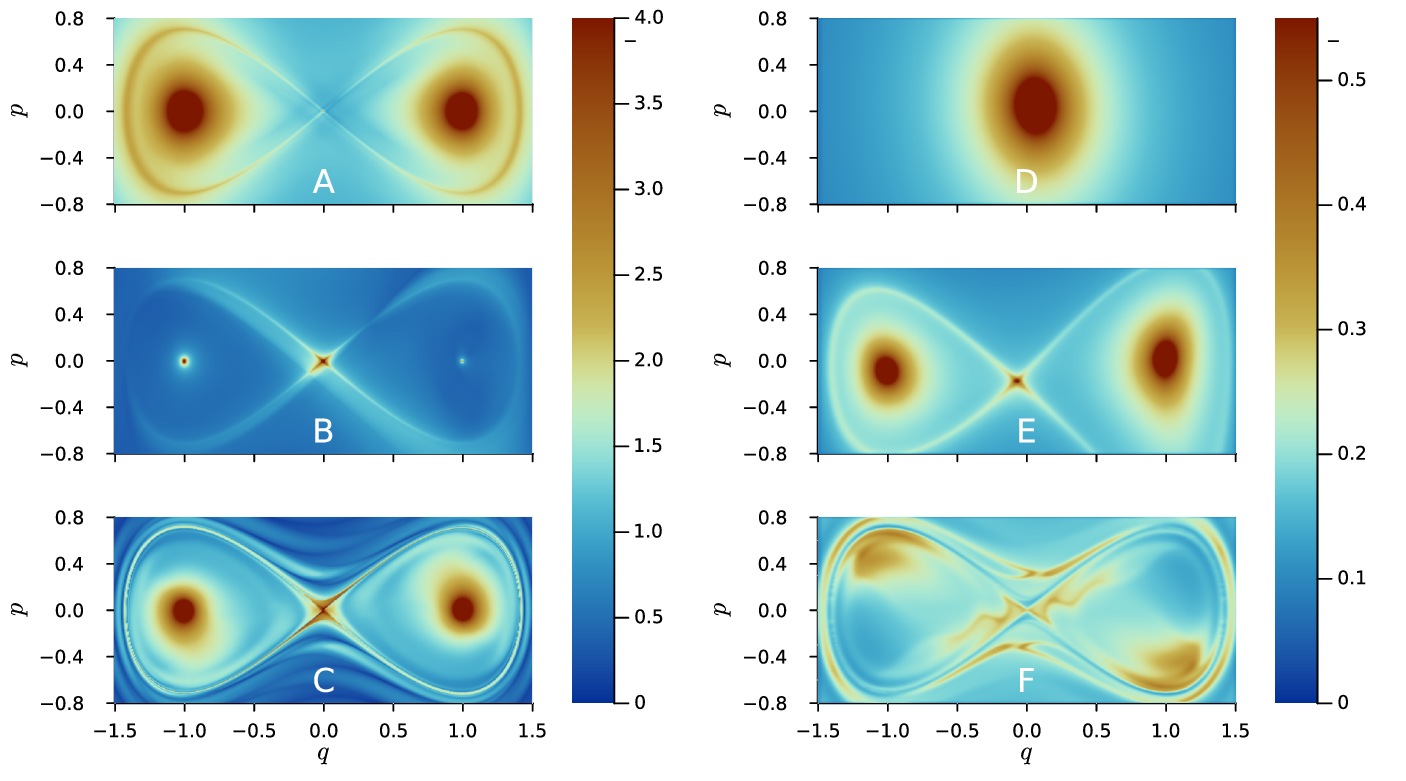}
    \caption[LD analysis of model failures with insufficient training data.]{Diagnostic visualization of model failures with insufficient training data. Left column (A--C): normalized prediction error distributions across phase space. Right column (D--F): corresponding LD-weighted PDFs. Top row (A,D): SympNet (100 trajectories) misses the figure-eight topology, learning instead a single center structure at $(0,0)$. Middle row (B,E): H\'enonNet (50 trajectories) produces a distorted homoclinic orbit with asymmetric lobes and incorrect curvature. Bottom row (C,F): RC (100 trajectories) captures the global homoclinic boundary but fails to reproduce the two center fixed points inside the orbit. LD analysis reveals specific structural errors that remain hidden in conventional error metrics.}
    \label{fig:LD_PDF_Fail_Duffing}
\end{figure}

To further illustrate model limitations, Fig.~\ref{fig:LD_PDF_Fail_Duffing} presents both normalized prediction error distributions (left column) and corresponding LD-weighted PDFs (right column) for representative cases with insufficient training data. Panel A shows the normalized prediction error for SympNet trained on 100 trajectories, with large discrepancies near the homoclinic orbit and center fixed points. The corresponding LD visualization (Panel D) reveals the cause: the model incorrectly learns a single center-focused structure at the origin $(0,0)$, missing the figure-eight topology. Panel B displays the error distribution for H\'enonNet trained on 50 trajectories, with errors concentrated along the homoclinic orbit. Its LD-weighted PDF (Panel E) confirms this, showing a distorted figure-eight with asymmetric lobes and incorrect curvature. For RC trained on 100 trajectories (Panels C and F), the global homoclinic boundary is qualitatively present, but elevated errors occur near the center fixed points, consistent with the missing structure in the corresponding LD representation.

This comparison illustrates the complementary value of normalized prediction errors and LDs in assessing NN performance for Hamiltonian systems. Conventional error metrics quantify local trajectory accuracy but do not necessarily reflect preservation of global phase-space structure. LDs, in contrast, expose how well models capture the fundamental phase-space geometry—including fixed points and homoclinic orbits—that governs system stability and long-term behavior.

\section{Numerical experiment: three-mode NLS}\label{subsec:three_modes}
We now extend the analysis to the three-mode truncation of the focusing Nonlinear Schr\"odinger (NLS) equation, which serves as a more challenging benchmark for assessing NN performance in modeling complex Hamiltonian dynamics. We begin by deriving the governing equations and describing the characteristic phase-space structure. After outlining dataset generation and NN configurations, the analysis proceeds in three stages: (i) short-time trajectory forecasting across distinct dynamical regimes, (ii) LD-based analysis of how accurately each NN reconstructs the global phase-space structure, including the homoclinic orbit geometry, and (iii) evaluation of the effects of both training dataset size and distribution, showing that strategic sampling across dynamically distinct regions strongly influences reconstruction fidelity.

\subsection{Governing equations}
The focusing NLS equation governs nonlinear wave dynamics in diverse physical contexts, including nonlinear optics and Bose–Einstein condensates \cite{zakharov_stability_1968,dystheOceanicRogueWaves2008,solli_optical_2007,bludov2009MatterRogue,mckerr2014FreakWaves,el-shafeay2023SuperRogue}:
\begin{equation}
    i\frac{\partial\psi}{\partial t} + \frac{1}{2}\frac{\partial^{2}\psi}{\partial\xi^{2}} + |\psi|^{2}\psi = 0, \label{eq:NLS_chap5}
\end{equation}
where $\psi(\xi,t)$ is the complex field variable depending on space $\xi$ and time $t$. This equation exhibits rich dynamical structures, including rogue-wave solutions that correspond to homoclinic orbits in phase space. While the NLS possesses an infinite-dimensional phase space, here we adopt a finite-dimensional approximation through a three-mode truncation, which enables systematic evaluation of NN performance. 

Unlike Cappellini et al.~\cite{cappellini1991ThirdorderThreewave}, who derived the truncation in amplitude–phase variables, we adopt a Cartesian representation that yields a four-degree-of-freedom Hamiltonian system. Substituting the ansatz
\begin{equation}
    \psi(\xi,t) = (q_0(t) + ip_0(t)) + \sqrt{2}\,(q_1(t) + ip_1(t))\cos(k\xi),
\end{equation}
with real-valued functions $q_0(t),p_0(t),q_1(t),p_1(t)$ and wavenumber $k$, retaining only the constant and first harmonic terms in the expansion, we obtain the coupled system
\begin{align}
    \dot{q}_0 &= -(q_0^2+p_0^2)p_0 - 2q_0q_1p_1 - p_0\left(q_1^2+3p_1^2\right),	\label{eq:2mode_q0}\\
    \dot{p}_0 &= (q_0^2+p_0^2)q_0 + 2p_0q_1p_1 + q_0\left(3q_1^2+p_1^2\right),\label{eq:2mode_p0}\\
    \dot{q}_1 &= -\tfrac{1}{2}k^{2}p_1-\tfrac{3}{2}p_1(q_1^2+p_1^2) - 2q_0p_0q_1 - p_1\left(q_0^2+3p_0^2\right),\label{eq:2mode_q1}\\
    \dot{p}_1 &= \tfrac{1}{2}k^{2}q_1+\tfrac{3}{2}q_1(q_1^2+p_1^2) + 2q_0p_0p_1 + q_1\left(3q_0^2+p_0^2\right).\label{eq:2mode_p1}
\end{align}

These equations follow from the non-separable Hamiltonian
\begin{equation}
    \begin{split}
        \mathcal{H}_{3\mathrm{NLS}} = -\Bigg[\frac{1}{4}\left(q_0^2+p_0^2\right)^2 + \frac{3}{8}\left(q_1^2+p_1^2\right)^2 + \frac{1}{4}k^2(q_1^2+p_1^2) + 2q_0p_0q_1p_1 \\
        +\frac{q_0^2}{2}(3q_1^2+p_1^2)+\frac{p_0^2}{2}(q_1^2+3p_1^2)\Bigg].
    \end{split}
\end{equation}
The non-separability of $\mathcal{H}_{3\mathrm{NLS}}$, particularly due to cross terms such as $q_0p_0q_1p_1$, introduces coupling between the mean field $(q_0,p_0)$ and the first Fourier mode $(q_1,p_1)$. This complexity makes the system a stringent test for symplectic-type NNs, which must preserve both energy conservation and symplectic structure. 

\subsubsection*{Dataset generation}
Training datasets were generated by numerically integrating Eqs.~\eqref{eq:2mode_q0}--\eqref{eq:2mode_p1} with initial conditions uniformly sampled as $\eta(0)\in (0,1)$ and $\phi(0) \in [-\pi,\pi)$. These polar coordinates $(\eta(0),\phi(0))$ were converted to Cartesian coordinates for NN input using
\begin{equation}
    q_0(0) = \sqrt{\eta(0)}\cos\phi(0),\quad  p_0(0) = -\sqrt{\eta(0)}\sin\phi(0), 
\end{equation}
\begin{equation}
    q_1(0) = \sqrt{1-\eta(0)},\quad  p_1(0) = 0, 
\end{equation}
with total power normalized to $P_0=1$. Trajectories were integrated for $k=0.95$, spanning 100 time units with output step size $\Delta t = 0.1$, yielding 1000 time points per trajectory. A dataset of 500 trajectories was split into training (80\%), validation (10\%), and testing (10\%) subsets.

\subsection{NN training and quality metrics}
For the three-mode NLS system, we employed architectures similar to those used for the Duffing equation, with adjusted hyperparameters to accommodate the higher-dimensional phase space. Key modifications include increasing GHNN’s number of learned Hamiltonians from 3 to 5 and raising RC’s sparsity parameter to 0.0075 (yielding approximately three connections per neuron). All symplectic architectures were trained on Cartesian coordinates $(q_0,p_0,q_1,p_1)$ using the Adam optimizer for 3000 epochs. Table~\ref{tab:nn_config_2mode} summarizes the architecture specifications. 

\begin{table}[htbp]
    \centering
    \small
    \begin{tabular}{lccccc}
        \hline
        NN type & \makecell{Learned \\ Hamiltonians} & Layers & \makecell{Neurons  \\ per layer} & \makecell{Trainable \\ parameters} & \makecell{FLOPs \\ at inference}\\
        \hline
        SympNet   & 10 & 1 & 50  & 4000  & 10000 \\
        H\'enonNet & 10 & 1 & 50  & 1010  & 10040 \\
        GHNN      & 5  & 2 & 15  & 10500 & 10800 \\
        RC        & -- & -- & 400 & 2000  & 9996  \\
        \hline
    \end{tabular}
    \caption{Hyperparameters of the NN architectures for the three-mode NLS system.}
    \label{tab:nn_config_2mode}
\end{table}

\begin{figure}
    \centering
    \includegraphics[width=0.9\linewidth]{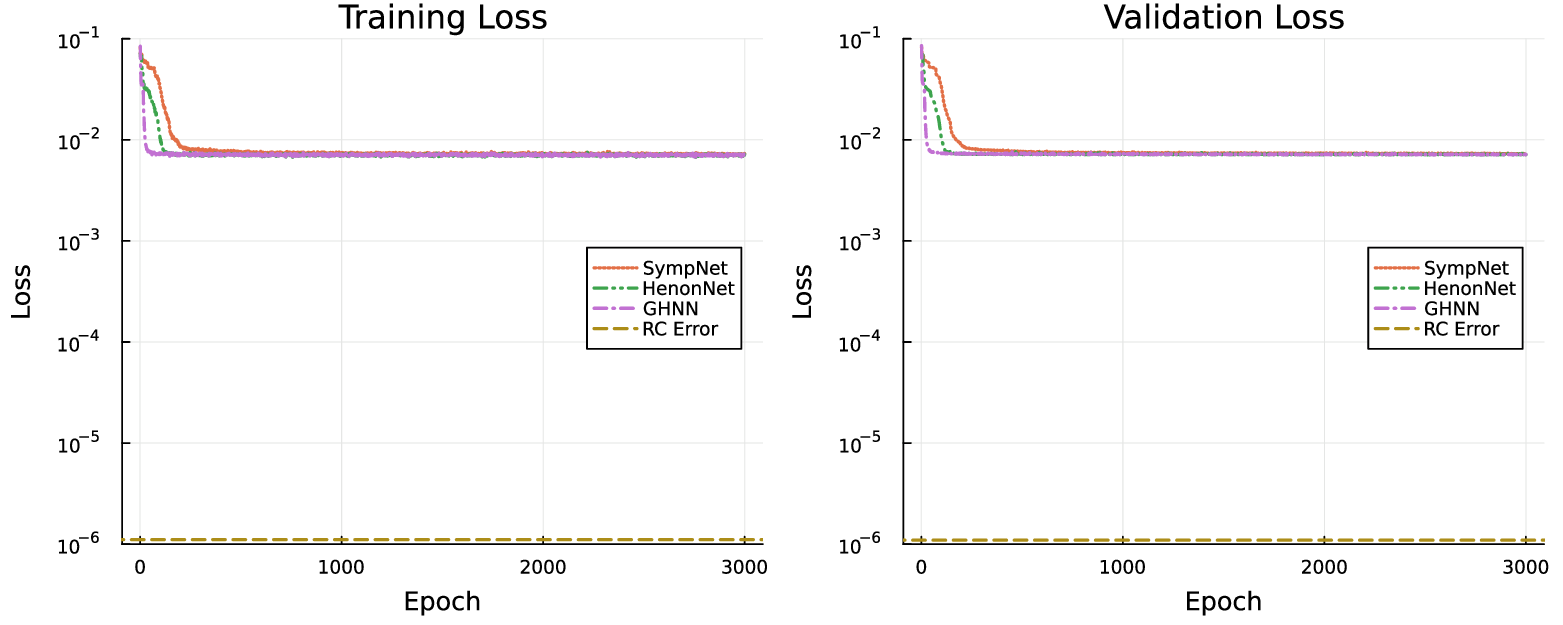}
    \caption{Training (left) and validation (right) loss for the NN models in the three-mode NLS case. 
The legend is similar with Fig.~\ref{fig:loss_compare_duffing}.}
    \label{fig:loss_compare_3mode}
\end{figure}

The training and validation loss (MSE) for the three-mode NLS case is shown in Fig.~\ref{fig:loss_compare_3mode}. For all symplectic-type architectures, the loss decreases during training and saturates at values slightly below $10^{-2}$. No further improvement is observed within the 3000 epochs, suggesting that the models have reached their effective convergence level under the current architecture and hyperparameter choices. For comparison, the normalized mean squared error (NMSE) obtained with Reservoir Computing (RC) is on the order of $10^{-6}$.

To analyze the homoclinic orbit geometry, we transform to polar coordinates:
\begin{equation}\label{eq:polar_subs}
     p_0+i q_0 = \zeta_0 e^{i\theta_{0}}, \quad  p_1+i q_1 = \zeta_1 e^{i\theta_{1}}, 
\end{equation} 
where $\zeta_j$ and $\theta_j$ represent the amplitudes and phases of each mode. The system conserves the total power $P_0$:
\begin{equation}
    \sum_{j=0}^1 q_j^2(t)+p_j^2(t) = P_0.
\end{equation}
This reduces the dynamics to the normalized amplitude $\eta = \zeta_0^2/P_0$ and the relative phase difference $\phi = \theta_1 - \theta_0$, with $\phi$ being the only dynamically relevant phase variable. 
In these reduced coordinates, the homoclinic orbit admits an explicit analytical representation (see \ref{App:PolarForm3Mode} and \cite{cappellini1991ThirdorderThreewave} for the detailed derivation), given by

\begin{equation}
    \cos(2\phi) = \frac{\tfrac{3}{4}\eta^{2} - \tfrac{1}{2}(1-k^2)\eta}{\eta(1-\eta)}. \label{eq:Hom_2D}
\end{equation}

To quantify model accuracy in this reduced coordinate system, we define the normalized prediction error in the $u_0 = (\eta(0),\phi(0))$ phase plane as
\begin{equation}
   \hat{e}(u_0) = \int_{-\tau}^{\tau} \left( \sum_{j=0}^1 |q_j(t)-\hat{q}_{j}(t)|^c+|p_j(t)-\hat{p}_{j}(t)|^c \right) dt,\quad  e(u_0) = \frac{\hat{e}(u_0)}{LD(u_0)}, \label{eq:error}
\end{equation}
where $\hat{q}_j(t)$ and $\hat{p}_j(t)$ denote the reference solution, $c=0.7$ is the scaling exponent, and $LD(u_0)$ is the corresponding LD. The LD is computed as
\begin{equation}
    LD(u_0) = \int_{-\tau}^{\tau} \left( \sum_{j=0}^1 |\dot{q}_j(t)|^c+|\dot{p}_j(t)|^c \right) dt.
\end{equation}
This normalization ensures error magnitudes remain comparable across regions with differing dynamical activity.

\subsection{Short-time prediction}
We now present representative short-term forward and backward forecasts for several initial conditions: near the homoclinic orbit, inside the homoclinic orbit, outside the orbit, and near the edges of phase space ($\eta \to 1$). 

\begin{figure}
    \centering
    \includegraphics[width=0.8\linewidth]{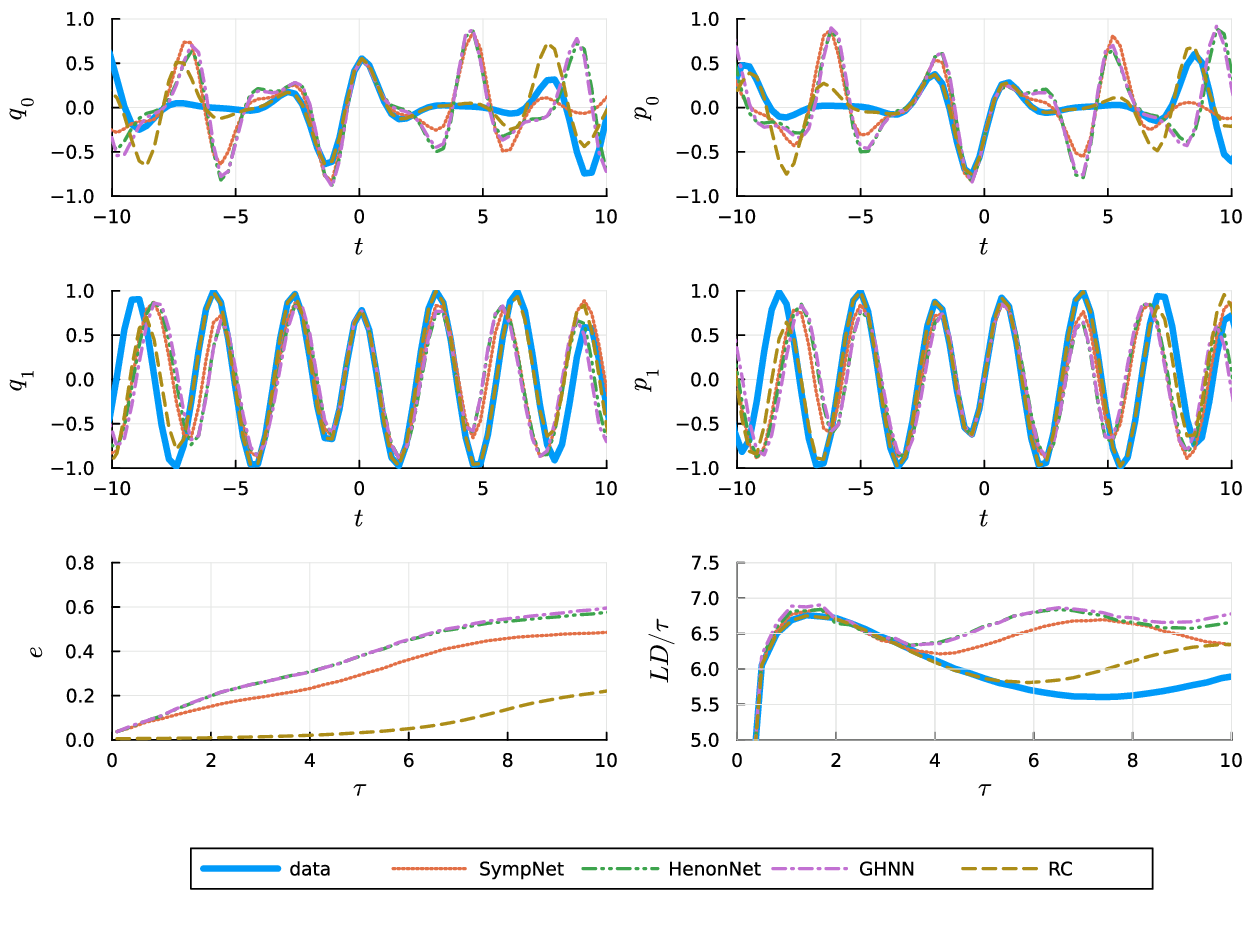}
    \caption[Short-term forecasts near the homoclinic orbit.]{Forward and backward forecasts of the three-mode NLS system, initialized near the homoclinic orbit at $u_0 = (\eta(0), \phi(0))=(0.4,0.57)$. Top rows: mode-0 dynamics ($q_0$, left; $p_0$, right) and mode-1 dynamics ($q_1$, left; $p_1$, right). Bottom row: pointwise prediction error (left) and normalized LD/$\tau$ (right) versus integration time $\tau$. Model comparisons follow the color/marker scheme of Fig.~\ref{fig:time_trace_Duff}. All NNs were trained on the 500-trajectory dataset.}
    \label{fig:time_trace_2NLS}
\end{figure}

Fig.~\ref{fig:time_trace_2NLS} shows predictions for $(\eta(0),\phi(0))=(0.4,0.57)$, corresponding to a point near the homoclinic orbit. The top two rows show mode-0 and mode-1 dynamics (position variables on the left, momentum variables on the right). The bottom row presents the pointwise error (left) and normalized LD (right) as functions of integration time $\tau$. All models initially agree well with the reference solution (blue solid line), but deviations grow with time, especially for mode 0. For mode 1, predictions exhibit systematic phase shifts, reflecting the difficulty of forecasting near the homoclinic orbit, where small perturbations induce large trajectory deviations.

\begin{figure}
    \centering
    \includegraphics[width=0.8\linewidth]{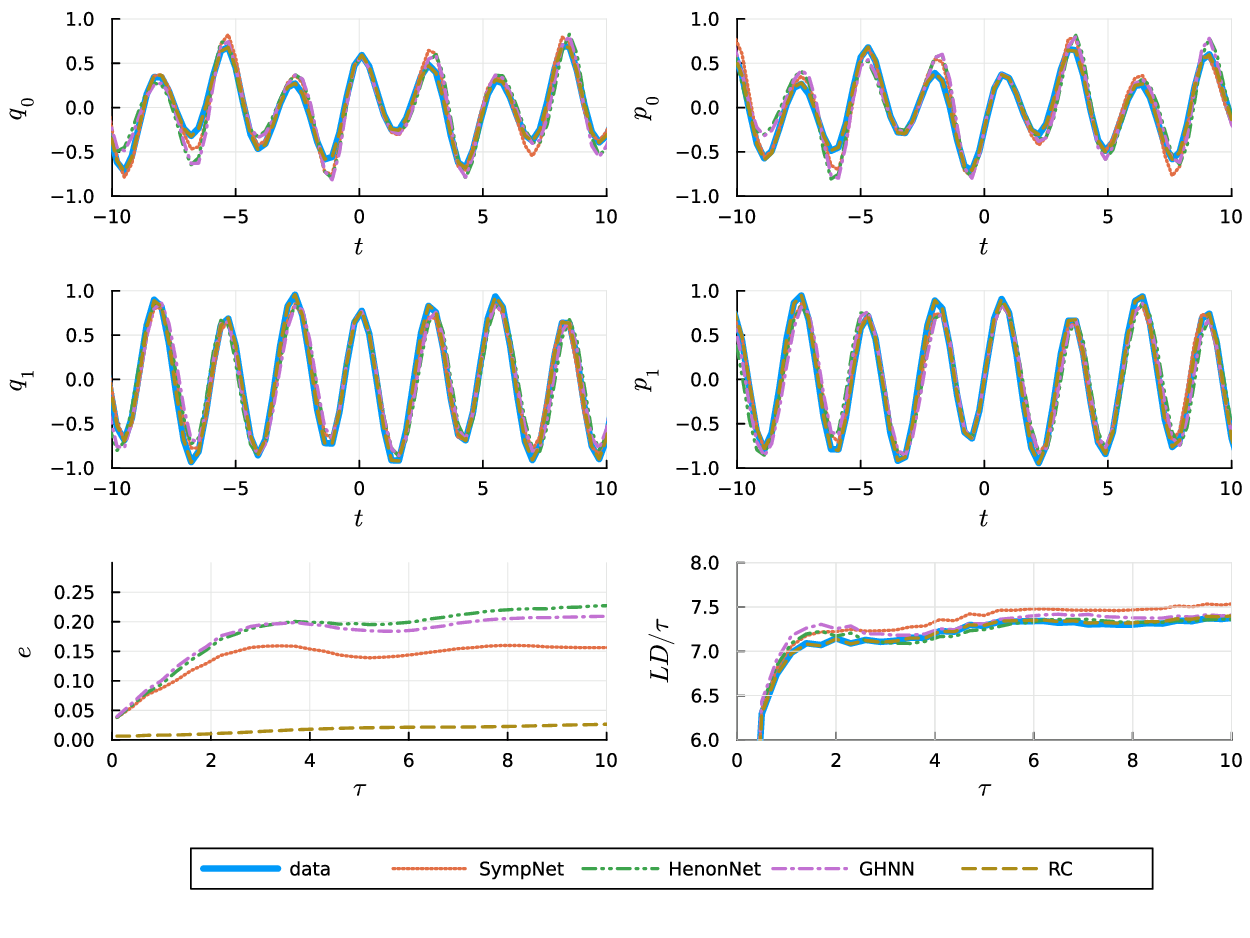}
    \caption[Short-term forecasts inside the homoclinic orbit.]{Same as Fig.~\ref{fig:time_trace_2NLS}, but for initial conditions $(\eta(0),\phi(0)) = (0.4,0.4)$ (inside the homoclinic orbit).}
    \label{fig:time_trace_2NLS_eta_0_4_phi_0_4}
\end{figure}

\begin{figure}
    \centering
    \includegraphics[width=0.8\linewidth]{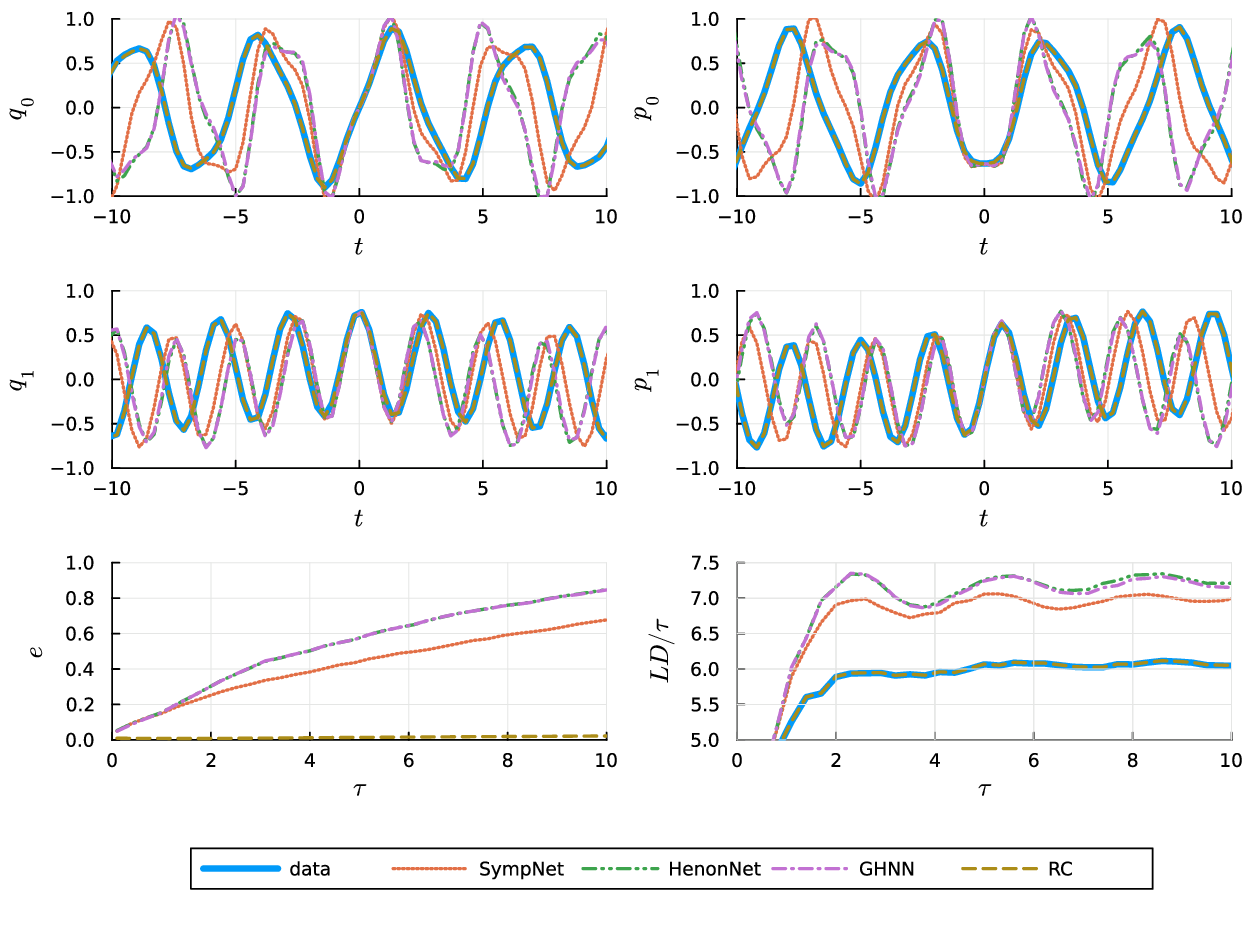}
    \caption[Short-term forecasts outside the homoclinic orbit.]{Same as Fig.~\ref{fig:time_trace_2NLS}, but for initial conditions $(\eta(0),\phi(0)) = (0.4,1.57)$ (outside the homoclinic orbit).}
    \label{fig:time_trace_2NLS_eta_0_4_phi_1_57}
\end{figure}

\begin{figure}
    \centering
    \includegraphics[width=0.8\linewidth]{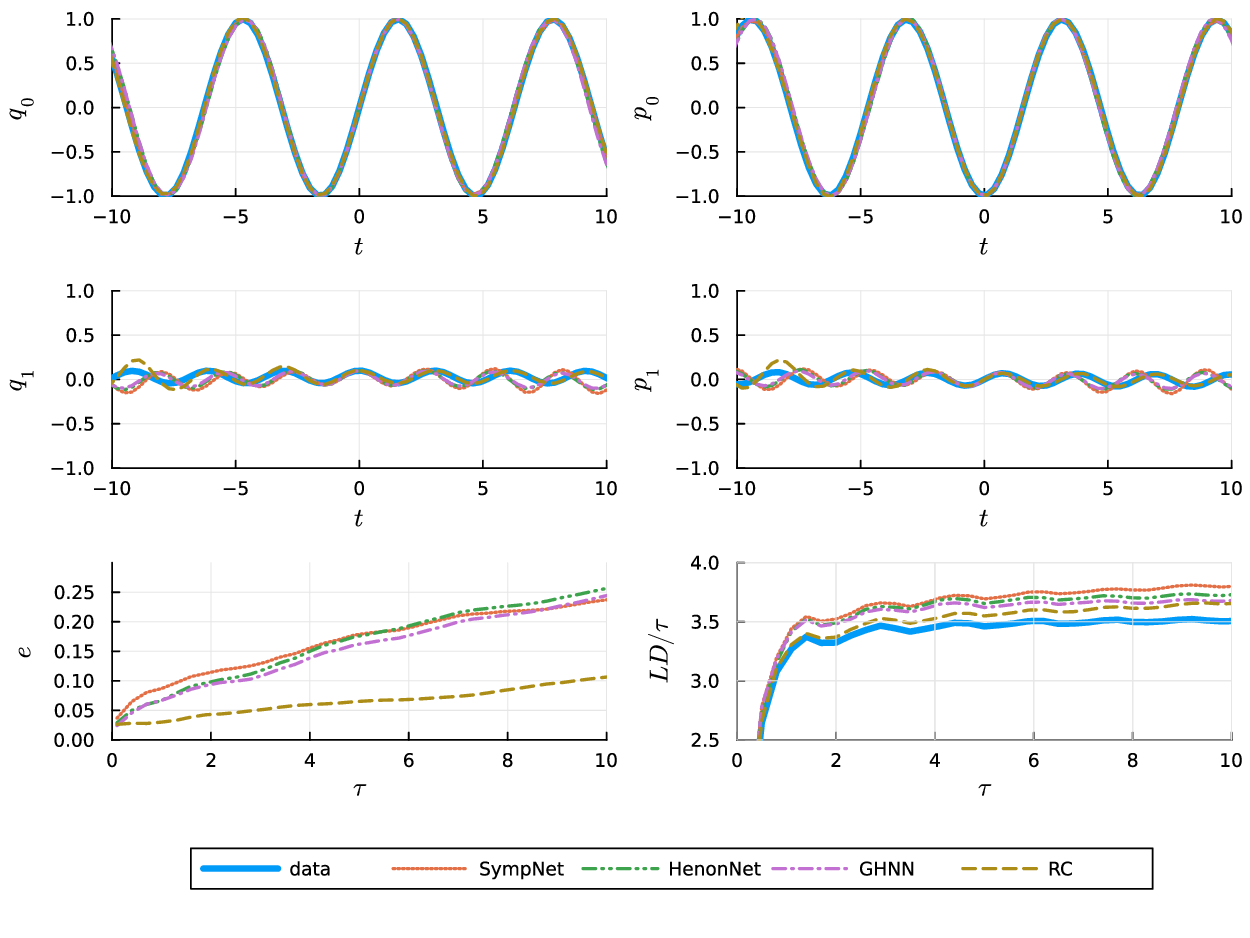}
    \caption[Short-term forecasts near the phase-space boundary.]{Same as Fig.~\ref{fig:time_trace_2NLS}, but for initial conditions $(\eta(0),\phi(0)) = (0.99,1.57)$ (near the phase-space boundary).}
    \label{fig:time_trace_2NLS_eta_0_99_phi_1_57}
\end{figure}

Figs.~\ref{fig:time_trace_2NLS_eta_0_4_phi_0_4}--\ref{fig:time_trace_2NLS_eta_0_99_phi_1_57} show analogous results for other initial conditions: $(\eta(0),\phi(0))=(0.4,0.4)$ (inside the homoclinic orbit), $(0.4,1.57)$ (outside the orbit), and $(0.99,1.57)$ (near the phase-space boundary). Inside the orbit, all models achieve good accuracy. RC consistently outperforms the other architectures outside the homoclinic orbit, including near the edges, maintaining accuracy for both forward and backward predictions. GHNN and H\'enonNet exhibit larger deviations, most evident in phase shifts, while SympNet shows significant errors in both amplitude and phase variables.

In contrast, the LDs (bottom right panel) provide a more robust metric for assessing long-term model performance. The normalized LD ($LD/\tau$) converges to a stable value as $\tau$ increases for trajectories away from the homoclinic orbit (Figs.~\ref{fig:time_trace_2NLS_eta_0_4_phi_0_4}--\ref{fig:time_trace_2NLS_eta_0_99_phi_1_57}), with its level sets corresponding to closed orbits in phase space. This convergence demonstrates the robustness of the LD-based approach and motivates the PDF-based LD methodology, which emphasizes the geometric features of the phase space.

\begin{figure}
    \centering
    \includegraphics[width=0.8\linewidth]{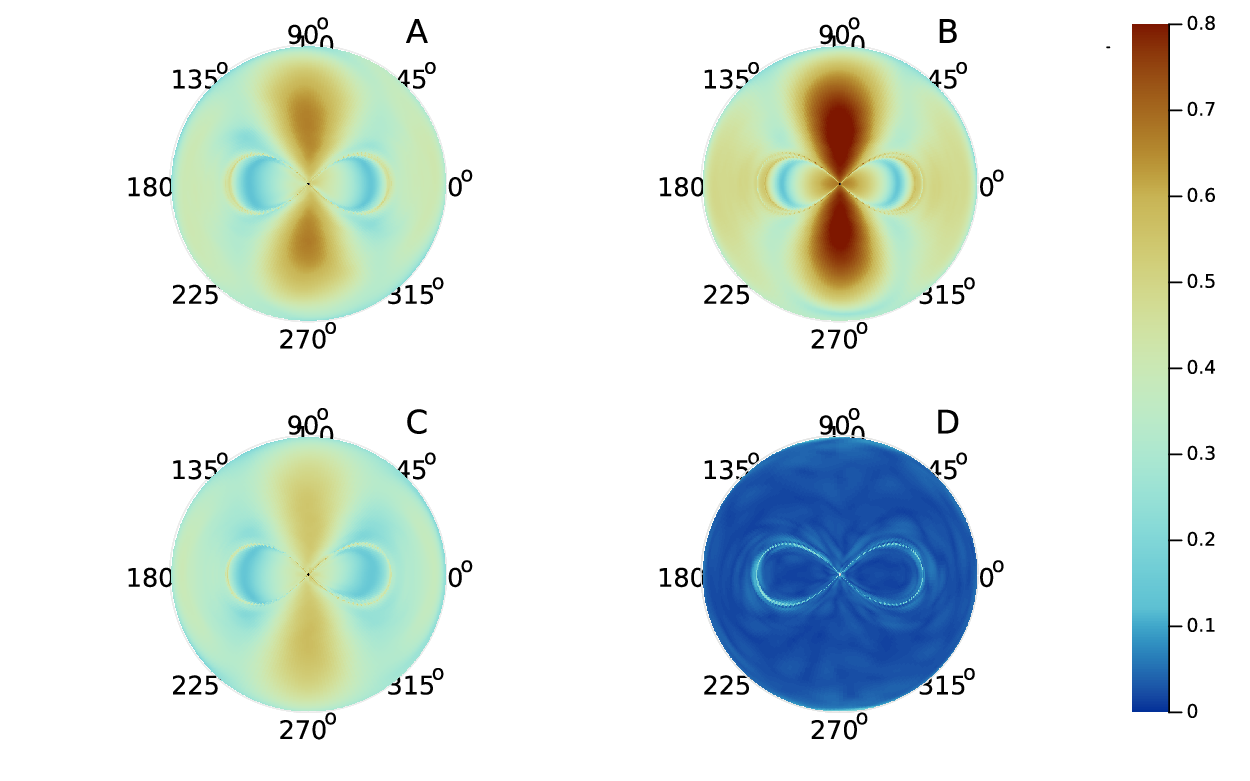}
    \caption[Error in the $(\eta,\phi)$ plane for NN models.]{Spatial distribution of the normalized prediction error in the $(\eta,\phi)$ plane for the three-mode NLS system, as defined in Eq.~\eqref{eq:error}. Errors are evaluated at $\tau=10$ and compared with the reference solution: (A) SympNet, (B) H\'enonNet, (C) GHNN, (D) RC.}
    \label{fig:LD_2mode_NRMSE}
\end{figure}

Fig.~\ref{fig:LD_2mode_NRMSE} shows the spatial distribution of the normalized prediction error in the $(\eta,\phi)$ plane at $\tau=10$. RC achieves the smallest overall error, although a noticeable error pattern traces the homoclinic orbit. Symplectic architectures (SympNet, H\'enonNet, and GHNN) display larger errors near $\phi=\pm\pi/2$, with H\'enonNet showing the largest deviations. These error distributions highlight the difficulty of accurately resolving dynamics in the vicinity of the homoclinic orbit.

\subsection{LD calculation}
We now show how LDs clarify the phase-space structures underlying the pointwise errors discussed above.  

LDs were computed on a uniform grid with $\eta \in [0,1]$ (500 points) and $\phi \in [-\pi,\pi]$ (200 points), using integration time $\tau=10$. The same procedure was applied to each model and to the reference system. The resulting LD values were then used to weight PDFs, highlighting dynamically significant regions of phase space.  

\begin{figure}
    \centering
    \includegraphics[width=0.9 \linewidth]{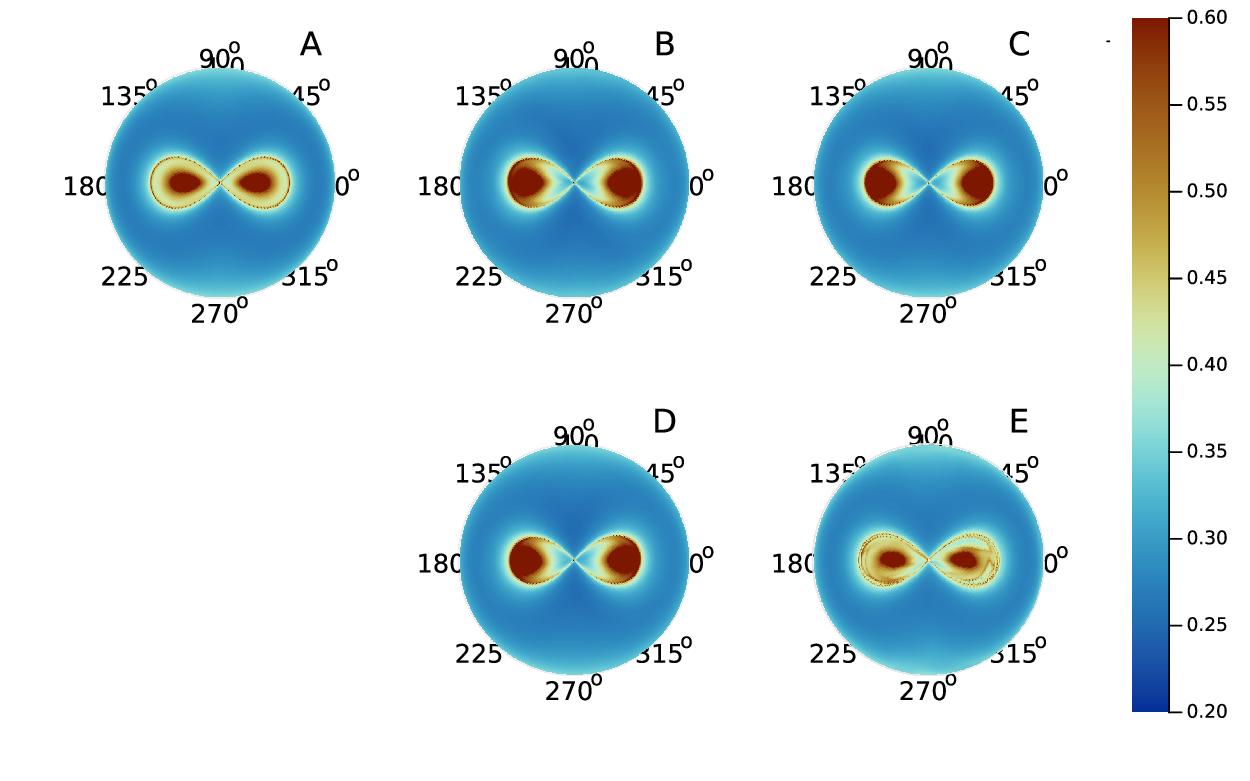}
    \caption[LD-weighted PDFs for the three-mode NLS system.]{LD-weighted PDFs for the three-mode NLS system in polar coordinates $(\eta,\phi)$. Panel A: reference solution from numerical integration. Panels B–E: predictions from SympNet, H\'enonNet, GHNN, and RC, respectively. RC most accurately reproduces the reference structure, resolving the homoclinic orbit and fixed points. SympNet, GHNN, and H\'enonNet display distorted density distributions, with displaced or missing fixed points inside the homoclinic orbit.}
    \label{fig:prob-2mode}
\end{figure}

Fig.~\ref{fig:prob-2mode} presents the LD-weighted PDFs. Panel A shows the reference PDF from numerical integration, while panels B–E show predictions from SympNet, H\'enonNet, GHNN, and RC, respectively. At first glance, all models reproduce the global figure-eight topology of the homoclinic orbit. However, closer inspection reveals critical differences. RC closely matches the reference structure, accurately resolving the homoclinic orbit, density inside the orbit, and the fixed points. In contrast, SympNet, GHNN, and H\'enonNet exhibit distorted density distributions within the homoclinic orbit, most notably misplaced fixed points. 

It is important to emphasize that regions of elevated pointwise reconstruction error (Figs.~\ref{fig:LD_2mode_NRMSE}) do not necessarily imply a complete loss of geometric structure. Small spatial displacements of invariant structures—such as slight shifts of fixed points or separatrices—can produce large local reconstruction errors while preserving the overall topology of the phase space. A particularly illustrative case is shown in Fig.~\ref{fig:time_trace_2NLS_eta_0_99_phi_1_57}, corresponding to $\eta=0.99$ and $\phi \approx \pi/2$. In this example, the variables $q_1$ and $p_1$ exhibit a shifted oscillation frequency while maintaining nearly the same amplitude. This results in substantial pointwise deviation over time, yet the associated LD values remain relatively similar because the global trajectory geometry is largely preserved.  Since the LD-weighted PDF reflects accumulated trajectory behavior rather than pointwise agreement, it can remain close to the reference even in regions where local errors are significant. The observed discrepancies therefore reflect geometric distortions rather than a total breakdown of the underlying dynamical organization.

\begin{figure}
    \centering
    \includegraphics[width=0.8\linewidth]{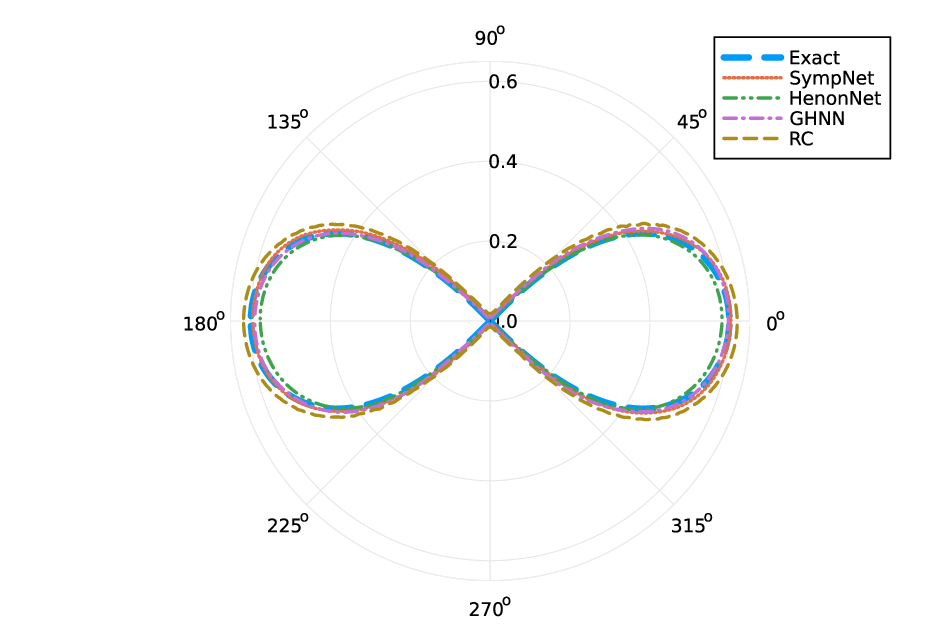}
    \caption[Homoclinic orbit predictions in polar coordinates $(\eta,\phi)$.]{Homoclinic orbit predictions in polar coordinates $(\eta,\phi)$ for NN models compared with the exact orbit (Eq.~\eqref{eq:Hom_2D}, blue dashed). Color scheme as in Fig.~\ref{fig:homoclinic_orbit}. RC reproduces the orbit with high fidelity, while the symplectic models preserve the overall figure-eight structure but underestimate its diameter and curvature. All models trained on 500 trajectories.}
    \label{fig:homoclinic_orbit_2mode}
\end{figure}

 Plotting the homoclinic orbit (Fig.~\ref{fig:homoclinic_orbit_2mode}) reveals important differences in how the models resolve its precise geometry. RC (orange dashed) provides the most accurate representation, closely matching the reference solution (blue dashed). The symplectic architectures capture the overall figure-eight structure but produce a contracted orbit with reduced diameter and altered curvature. This geometric compression indicates that while these models preserve the homoclinic structure topologically, they struggle to reproduce its quantitative features. This limitation likely arises from the non-separable nature of the three-mode NLS Hamiltonian, which challenges their parameterization.  

Quantitative analysis using the KL divergence (Table~\ref{tab:kl_3mode}) confirms these visual observations: RC achieves values up to two orders of magnitude lower than those of the symplectic models across the LD-weighted PDFs in Fig.~\ref{fig:prob-2mode}. This highlights RC’s unexpected advantage in capturing complex phase-space geometries despite lacking explicit Hamiltonian priors.  

\subsection{Effect of training dataset}
We next investigate the effect of training dataset size and distribution on model performance. In addition to varying the number of trajectories $N \in \{125, 250, 500, 1000\}$, we compare three training distributions:
\begin{itemize}
    \item Uniformly distributed inside and outside the homoclinic orbit ($\eta(0)\in[0,1],\;\phi(0)\in[-\pi,\pi]$),
    \item Uniformly distributed inside the homoclinic orbit only,
    \item Uniformly distributed outside the homoclinic orbit only.
\end{itemize}

\begin{table}[htbp!]
\centering
   \scriptsize
\begin{tabular}{llcccc}
    \hline
\multirow{2}{*}{Training distribution} & \multirow{2}{*}{NN model} & \multicolumn{4}{c}{Number of training trajectories $N$} \\
\cline{3-6} 
 &  & 125  & 250 & 500 & 1000 \\
\hline
\multirow{4}{*}{Uniform (inside + outside)} & SympNet   & 2.14e-03 & 2.81e-03 & 2.36e-03 & 3.39e-03 \\
 & H\'enonNet & 3.59e-03 & 3.60e-03 & 3.73e-03 & 1.18e-02 \\
 & GHNN       & 3.90e-03 & 1.87e-03 & 3.47e-03 & 1.86e-03 \\
 & RC         & \textbf{1.72e-04} & \textbf{8.11e-05} & \textbf{1.04e-05} & \textbf{3.60e-04} \\
\hline
\multirow{4}{*}{Inside only} & SympNet   & \textbf{3.35e-03} & 5.07e-03 & \textbf{4.45e-03} & \textbf{4.76e-03} \\
 & H\'enonNet & 3.93e-03 & 6.19e-03 & 4.61e-03 & 1.18e-02 \\
 & GHNN       & 3.40e-03 & 4.20e-03 & 4.58e-03 & 5.21e-03 \\
 & RC         & 8.89e-03 & \textbf{3.56e-03} & 4.76e-03 & 6.56e-03 \\
\hline
\multirow{4}{*}{Outside only} & SympNet   & 5.31e-03 & 2.95e-03 & 2.33e-03 & 2.38e-03 \\
 & H\'enonNet & 3.85e-03 & 4.16e-03 & 3.90e-03 & 1.18e-02 \\
 & GHNN       & 2.50e-03 & 2.56e-03 & 2.49e-03 & 2.26e-03 \\
 & RC         & \textbf{1.13e-04} & \textbf{7.41e-05} & \textbf{6.38e-05} & \textbf{1.36e-04} \\
\hline
\end{tabular}
\caption[KL divergence for the three-mode NLS under different training conditions.]{KL divergence between model-predicted and reference LD-weighted PDFs for the three-mode NLS, evaluated across training dataset sizes $N=\{125,250,500,1000\}$. Lower values indicate better reconstruction of the global phase-space structure. Bold entries mark the best-performing model for each condition. RC achieves superior performance (by 1–2 orders of magnitude) with uniform and outside-only training, while all models perform similarly when trained exclusively inside the homoclinic orbit.}
\label{tab:kl_3mode}
\end{table}

Table~\ref{tab:kl_3mode} highlights several important trends. First, training distribution strongly affects phase-space reconstruction, often more than increasing the dataset size. With uniformly distributed data, RC consistently outperforms the symplectic architectures by 1–2 orders of magnitude, reaching KL divergence as low as $1.04\!\times\!10^{-5}$ at $N=500$. This advantage persists under outside-only training.  

By contrast, all models—including RC—struggle when trained only on inside-homoclinic trajectories. Under these conditions, performance gaps narrow, and SympNet occasionally surpasses RC. This indicates that quasi-periodic trajectories confined inside the homoclinic orbit provide insufficient information to reconstruct the global phase-space structure, especially the homoclinic boundary.   

\begin{figure}
    \centering
    \includegraphics[width=\linewidth]{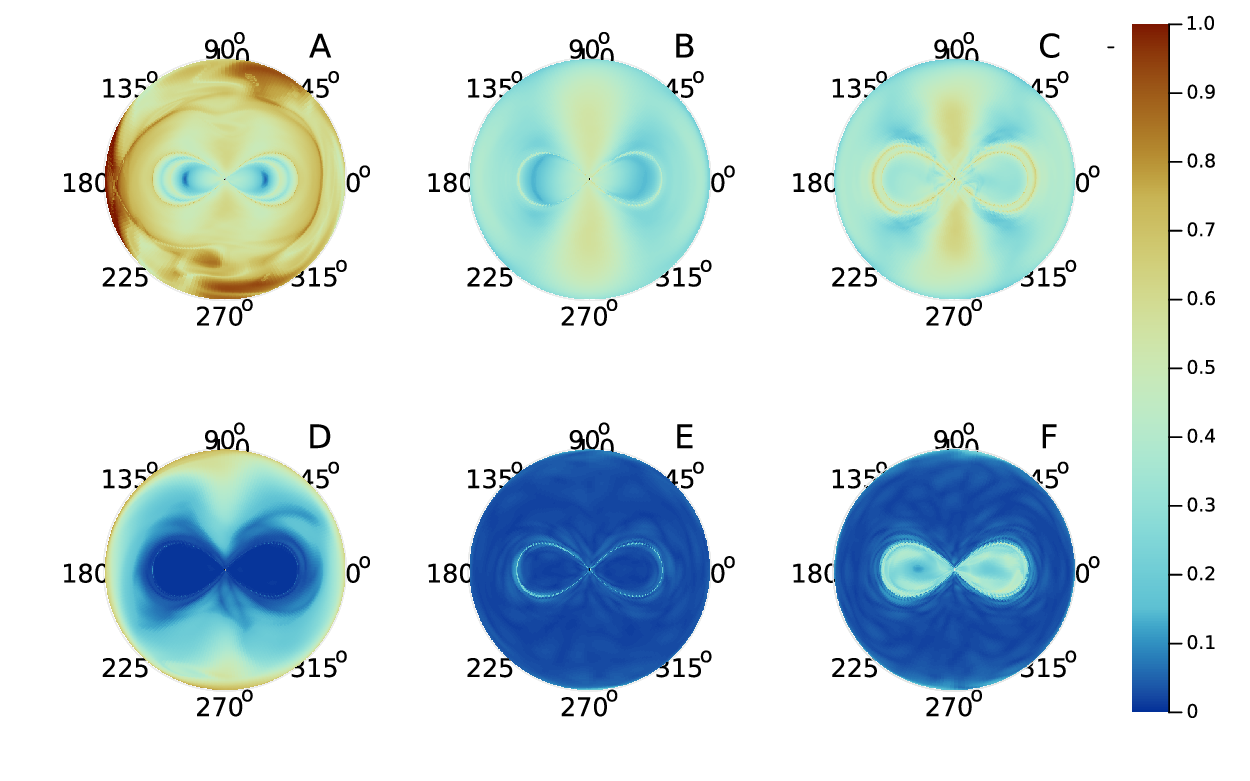}
    \caption[Error distributions under different training distributions.]{Normalized prediction error in the $(\eta,\phi)$ phase plane for GHNN (top row) and RC (bottom row) trained on different data distributions: inside-homoclinic only (left), uniform (middle), and outside-homoclinic only (right). Training with biased data yields complementary error patterns: inside-only (outside-only) training produces large errors in outside (inside) regions. Even with uniform data, all models show increased errors along the homoclinic boundary, though RC maintains substantially lower errors overall.}
    \label{fig:LD_2mode_NRMSE_in_out}
\end{figure}

\begin{figure}
    \centering
    \includegraphics[width=0.9\linewidth]{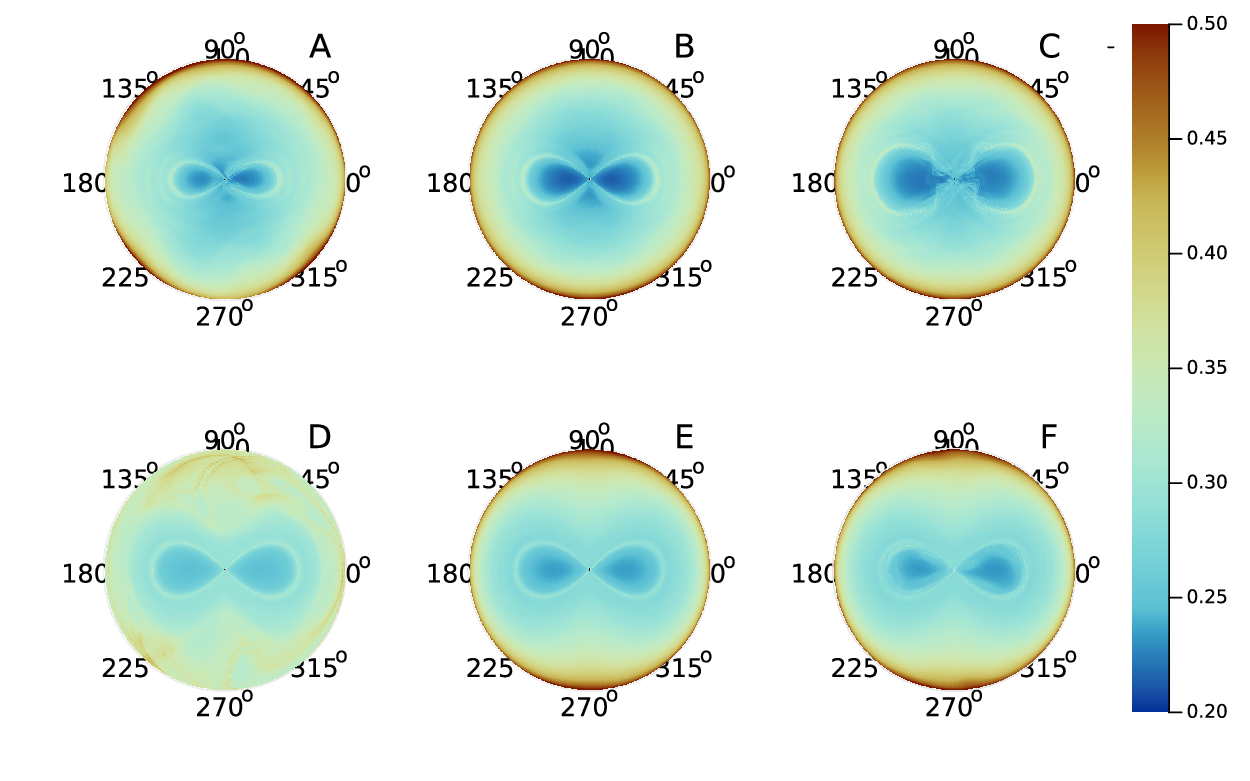}
    \caption[LD-weighted PDFs under different training distributions.]{LD-weighted PDFs for cases correspond to Fig.~\ref{fig:LD_2mode_NRMSE_in_out}. Training distribution dramatically affects orbit reconstruction: inside-only training produces contracted orbits, while outside-only training yields blurred orbits near the saddle point at $\phi=0$. RC better preserves global topology across all cases compared to the symplectic architectures.}
    \label{fig:LD_2mode_in_out}
\end{figure}

To visualize these effects, Figs. \ref{fig:LD_2mode_NRMSE_in_out} and \ref{fig:LD_2mode_in_out} present the normalized prediction error and LD-weighted PDFs, respectively, for GHNN (top row) and RC (bottom row) under different training distributions (inside-homoclinic, uniform, and outside-homoclinic, from left to right). These visualizations reveal distinct failure modes for each model and training condition. 

For GHNN, training exclusively inside the homoclinic orbit produces a severely contracted homoclinic boundary. Conversely, training solely on outside trajectories yields a distorted homoclinic structure with irregular curvature near the saddle point at $\phi=0$. For RC, inside-only training results in poor resolution of the high-density region for $\eta \to 1$, while outside-only training produces a blurred representation of the homoclinic orbit and interior structures, though the global topology remains qualitatively correct.

These visualizations confirm a fundamental challenge in Hamiltonian system learning: model generalization beyond the training data distribution is limited by the system's rich phase space structure, where different regions exhibit qualitatively distinct dynamics. This suggests that strategic sampling across diverse dynamical regimes, particularly near critical structures like homoclinic orbits, may be more important than simply increasing training data volume. The superior performance of RC under various training conditions further indicates that architectural flexibility may outweigh explicit physical constraints when learning complex phase space geometries.

\section{Sensitivity analysis of LD-weighted PDF Framework}\label{sec:sensitivity_analysis}

In this section, we investigate the sensitivity of the LD-weighted PDF framework to key parameters, including the integration time $\tau$, the exponent $c$ in the LD definition, and the choice of weighting function $g$ in the LD-weighted PDF construction. Understanding how these parameters influence the detection and quantification of dynamical structures is crucial for establishing robust model evaluation protocols. We focus our discussion on the Duffing oscillator system, using the 200-trajectory case as a representative example. The baseline parameters employed in the main experiments are: $\tau=4$, $c=0.7$, $g(x) = 1/x$.

\begin{figure}[ptb]
    \centering
    \includegraphics[width=0.9\linewidth]{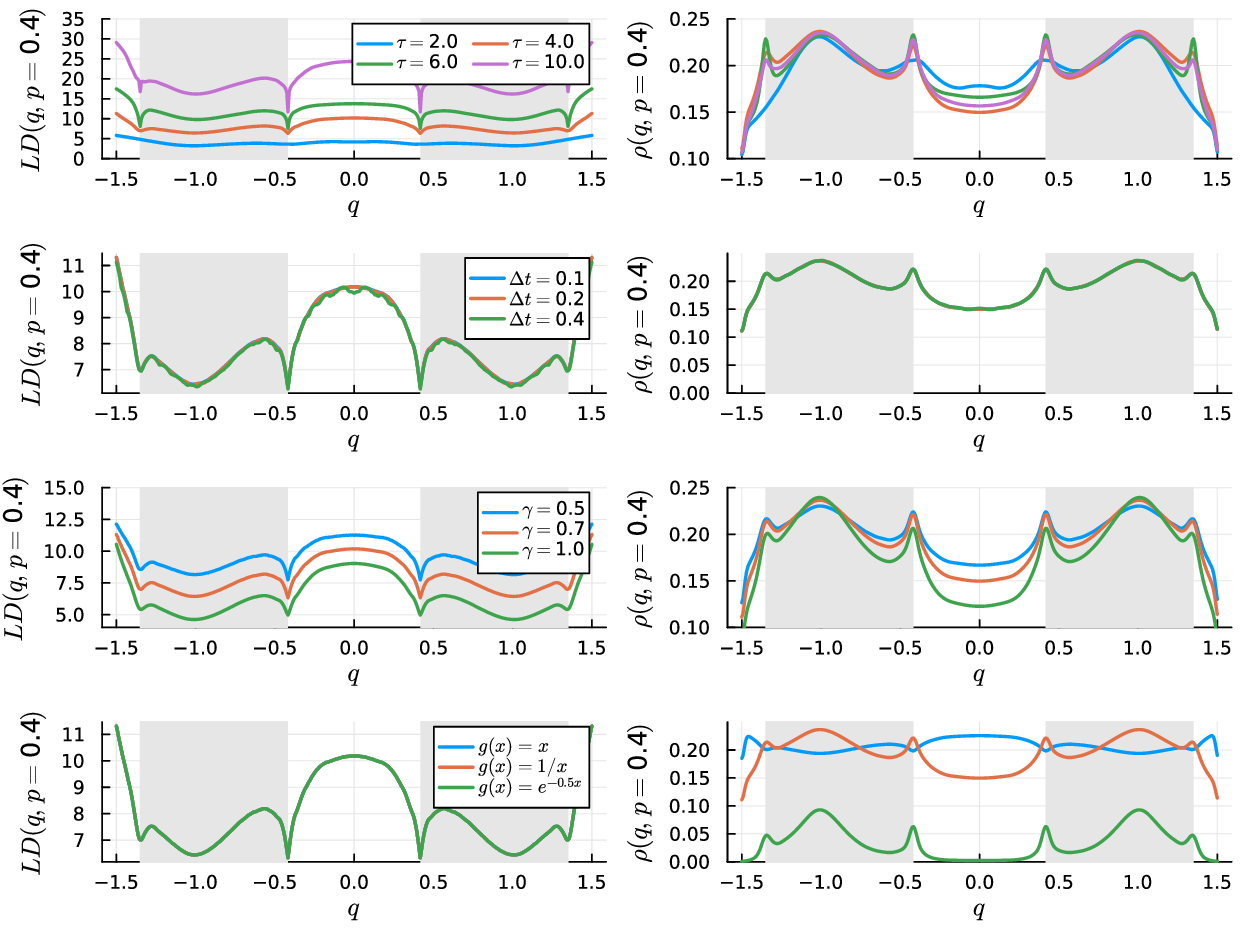}
    \caption[Sensitivity analysis of LD-weighted PDF framework parameters for the Duffing oscillator system.]{ Panels show cross-sections of computed LD values (left column) and corresponding LD-weighted PDFs $\rho$ at $p=0.4$ (right column) for varying integration time $\tau$ (top row), LD exponent $c$ (middle row), and weighting function $g$ (bottom row). The baseline parameters are $\tau=4$, $c=0.7$, and $g(x) = 1/x$ (shown as red curves). Gray shaded regions indicate areas inside the homoclinic orbit. Different parameter choices systematically affect the relative importance assigned to various phase-space regions.}
     \label{fig:LD_params}
\end{figure}

\begin{figure}[ptb]
    \centering
    \includegraphics[width=0.9\linewidth]{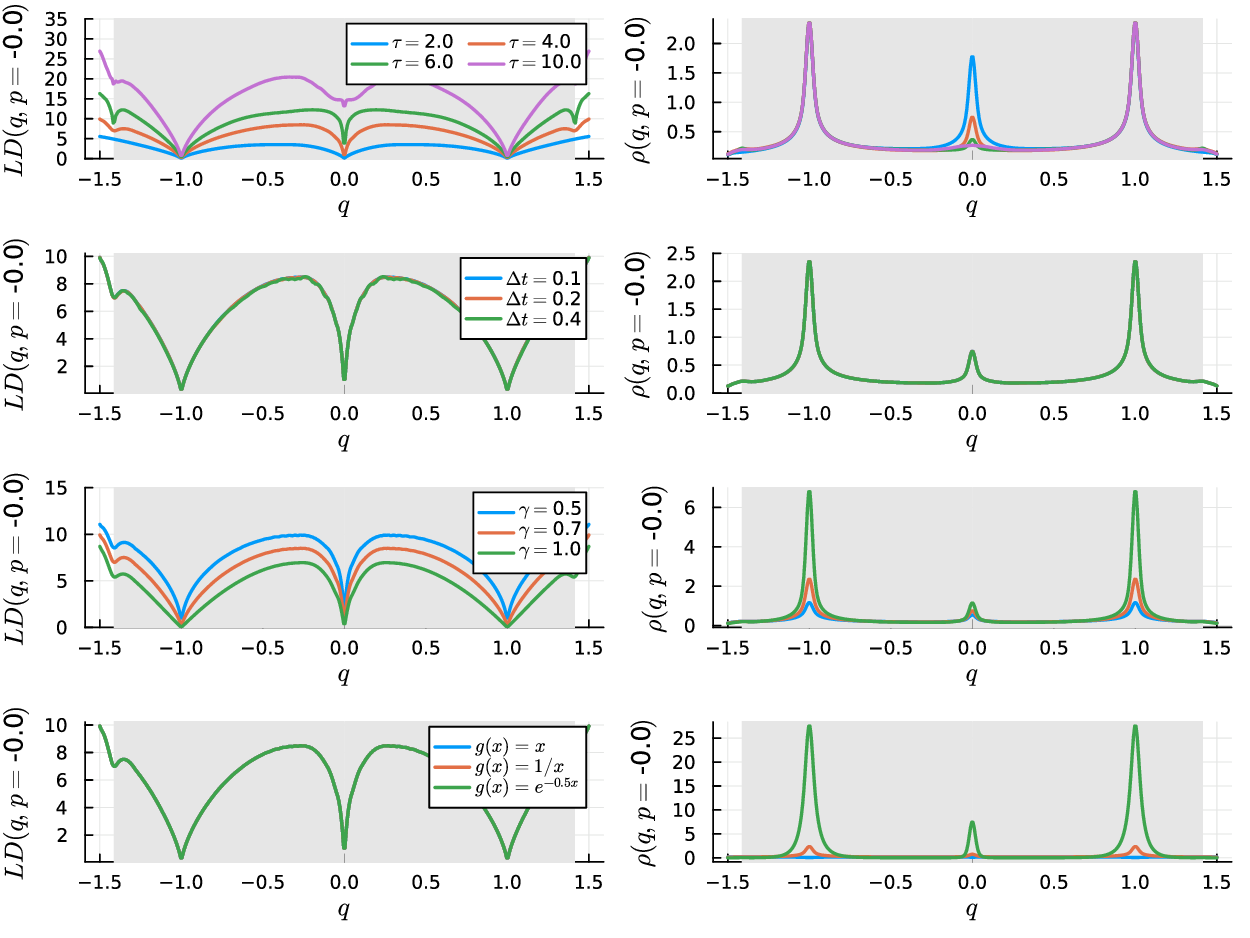}
    \caption[Similar to Fig.~\ref{fig:LD_params}, but for $p=0$.]{Similar to Fig.~\ref{fig:LD_params}, but for $p=0$.}

     \label{fig:LD_params_p0}
\end{figure}

Different parameter choices alter the emphasis placed on different phase-space regions, thereby influencing the computed LD values and resulting PDFs. To make this point apparent, we plot the LD values and resulting PDFs of the reference Duffing equation in two representative cross-sections, namely $p=0.4$ (Figs.~\ref{fig:LD_params}) and $p=0$ (Figs.~\ref{fig:LD_params_p0}). This highlights the importance of carefully selecting these parameters based on the specific dynamical features of interest in a given system.

This investigation demonstrates that while specific parameter choices can affect KL divergence values in Table~\ref{tab:kl_duffing}, the overall ranking of model performance remains consistent across different parameter settings. We therefore summarize the results using tables that compare relative rankings across multiple distance metrics and parameter configurations, with rankings indicated in parentheses (1 = best performance, higher numbers indicate relatively poorer performance). Superscripts * in the parameter column entries indicate the baseline setting for each parameter.

\begin{table}
    \centering
\begin{tabular}{llccccc}
\hline
Section & Parameter & SympNet & H\'enonNet & GHNN & RC \\
\hline
$\tau$ & 2.0 & 1.28e-03 (2) & 9.11e-03 (4) & 2.95e-03 (3) & 3.09e-04 (1) \\
  & 4.0* & 1.32e-03 (2) & 8.03e-03 (4) & 2.75e-03 (3) & 6.20e-05 (1) \\
 & 6.0 & 1.74e-03 (2) & 8.50e-03 (4) & 3.24e-03 (3) & 3.22e-05 (1) \\
 & 10.0 & 1.30e-03 (2) & 6.90e-03 (4) & 2.55e-03 (3) & 2.51e-05 (1) \\
 \hline
$\Delta t$ & 0.1* & 1.32e-03 (2) & 8.03e-03 (4) & 2.75e-03 (3) & 6.20e-05 (1) \\
 & 0.2 & 1.31e-03 (2) & 8.02e-03 (4) & 2.74e-03 (3) & 6.25e-05 (1) \\
& 0.4 & 1.31e-03 (2) & 8.01e-03 (4) & 2.74e-03 (3) & 8.53e-05 (1) \\
\hline
$c$ & 0.5 & 5.44e-04 (2) & 3.20e-03 (4) & 1.11e-03 (3) & 3.46e-05 (1) \\
  & 0.7* & 1.32e-03 (2) & 8.03e-03 (4) & 2.75e-03 (3) & 6.20e-05 (1) \\
  & 1.0 & 4.47e-03 (2) & 3.32e-02 (4) & 9.58e-03 (3) & 1.14e-04 (1) \\
\hline
$g(x)$ & $x$ & 6.17e-05 (2) & 6.35e-04 (4) & 1.19e-04 (3) & 5.38e-06 (1) \\
& $1/x$* & 1.32e-03 (2) & 8.03e-03 (4) & 2.75e-03 (3) & 6.20e-05 (1) \\
 & $\exp(-0.5x)$ & 2.81e-02 (2) & 1.52e-01 (4) & 6.15e-02 (3) & 9.53e-04 (1) \\
\hline

\end{tabular}
    \caption[Sensitivity analysis of LD-weighted PDF framework for the Duffing oscillator.]{Sensitivity analysis of the LD-weighted PDF framework for the Duffing oscillator system using 200 training trajectories. The table summarizes KL divergence values between model-predicted and reference LD-weighted PDFs under variations in integration time $\tau$, time discretization $\Delta t$, exponent $c$ in the LD definition, choice of weighting function $g$. Rankings (in parentheses) indicate relative model performance, with 1 being the best. RC consistently outperforms symplectic architectures across all parameter settings, demonstrating the robustness of our comparative framework.}
    \label{tab:tau_sensitivity}
\end{table}
\subsection{Effect of integration time}
For the integration time $\tau$, previous studies \cite{montes2021LagrangianDescriptors} have shown that in regular regions of phase space, LD values converge rapidly with increasing $\tau$, (see also right columns of Figs.~\ref{fig:time_trace_Duff}- \ref{fig:time_trace_Duff_plus} and Appendix \ref{App:LD_Ham} for the harmonic oscillator). While in our experiments, most of the regions are regular, while the only chaotic region is close to the homoclinic orbit for the Duffing oscillator. We vary $\tau$ in the range $\{2,4,6,8\}$ and compute the KL divergence between the LD-weighted PDFs for each model and the reference solution. 

The top panels of Figs.~\ref{fig:LD_params} and \ref{fig:LD_params_p0} show cross-sections of LD values (left) and the corresponding LD-weighted PDF $\rho$ (right) at $p=0.4$ and $p=0.$ respectively for the reference solution of the Duffing oscillator. We observe that while overall values of $\rho$ are quantitatively similar, the singular features become more pronounced for larger $\tau$. Notably, for $\tau=2$, the singular features are not captured in either the LD field or the corresponding $\rho$, indicating insufficient integration time to resolve the homoclinic structure. Similar trends are observed at $p=0$ (Fig.~\ref{fig:LD_params_p0}), where $\rho$ values remain nearly identical accross different integration times, including near the center fixed points at $q=\pm1$. However, a notable exception occurs at the saddle point ($q=0$), where smaller values of $\tau$ produce larger peaks in the PDF. This behavior reflects the sensitivity of the LD computation to integration time in the vicinity of hyperbolic fixed points. 

The associated KL divergence values in Table~\ref{tab:tau_sensitivity} demonstrate that while absolute KL values vary with $\tau$, the relative rankings of model performance remain remarkably stable across all integration times. RC consistently outperforms symplectic architectures regardless of the chosen integration horizon, confirming the robustness of our comparative framework

\subsection{Effect of time discretization}
In this section, we examine the effect of time discretization on the computation of the Lagrangian Descriptor (LD). Importantly, the learned NN models remain unchanged; only the time step $\Delta t$ used in the LD calculation is varied. In addition to the benchmark value $\Delta t = 0.1$, we consider $\Delta t \in {0.2, 0.4}$ and compute the Kullback–Leibler (KL) divergence between the resulting LD-weighted PDFs for each model and the corresponding reference solution. As reported in Table~\ref{tab:tau_sensitivity}, only minor variations in the KL divergence are observed, indicating that the LD-based comparison is robust with respect to moderate changes in time discretization.

The second-row panels of Figs.~\ref{fig:LD_params} and \ref{fig:LD_params_p0} display cross-sections of the LD field (left) and the corresponding LD-weighted PDF $\rho$ (right) at $p=0.4$ and $p=0.0$, respectively, for the reference Duffing solution. The overall structure remains essentially unchanged as $\Delta t$ varies. Only small oscillatory artifacts appear for larger values of $\Delta t$, and these effects do not alter the qualitative phase-space structure.

\subsection{Effect of LD exponent}
The exponent $c$ in the LD definition (Eq.~\eqref{eq:LD_arc_length}) controls the singularity strength of the LD field at invariant manifolds. Previous theoretical studies \cite{lopesino2017TheoreticalFramework} have established that singular features appear for $0 < c \leq 1$, with smaller values of $c$ producing sharper singularities. We vary $c$ in the range $\{0.5, 0.7, 1.0\}$ to assess how this parameter affects the detection and quantification of homoclinic structures.

The third-row panels of Figs.~\ref{fig:LD_params} and \ref{fig:LD_params_p0} show cross-sections of LD values and corresponding LD-weighted PDFs for different values of $c$. We observe that smaller values of $c$ (e.g., $c=0.5$) produce more pronounced singular features in the LD field, with sharper peaks near the homoclinic orbit and correspondingly higher density concentrations in the PDF at these locations. Conversely, larger values of $c$ (e.g., $c=1.0$) yield smoother LD fields with more diffuse PDF distributions, potentially reducing the contrast between regular and hyperbolic regions.

In all cases, the center fixed points exhibit larger $\rho$ values compared to the homoclinic orbit region. However, the contrast between these two dynamically distinct regions becomes less pronounced for smaller values of $c$, indicating that the choice of exponent affects the relative emphasis placed on different phase-space structures. This behavior is particularly evident at $p=0$ (Fig.~\ref{fig:LD_params_p0}), where the PDF peaks are concentrated at the center fixed points at $q=\pm1$. As in previous cases, the KL divergence values in Table~\ref{tab:tau_sensitivity} confirm that while absolute KL values vary with $c$, the relative rankings of model performance remain consistent across all tested values. 

\subsection{Effect of weighting function}
The choice of weighting function $g$ in the PDF definition (Eq.~\eqref{eq:PDF_LD}) fundamentally determines how LD values are translated into probability densities, affecting which phase-space regions receive emphasis in the analysis. We compare three different weighting functions to assess this sensitivity: $g(x) = x$, $g(x) = 1/x$ , and $g(x) = \exp(-0.5x)$.

The bottom panels of Figs.~\ref{fig:LD_params} and \ref{fig:LD_params_p0} show cross-sections of LD values and corresponding LD-weighted PDFs for these different weighting functions. The choice of $g$ dramatically alters the resulting PDF structure. With $g(x) = x$, the PDF emphasizes high LD regions, resulting in troughs near the homoclinic orbit and peaks outside the homoclinic region. Conversely, $g(x) = 1/x$ produces pronounced peaks in the PDF at the homoclinic orbit, highlighting its dynamical significance. The exponential weighting $g(x) = \exp(-0.5x)$ exhibits behavior similar to inverse weighting but with more pronounced selectivity. It generates the sharpest peaks at the lowest LD regions, particularly at the center fixed points $(q,p) = (\pm1,0)$, while yielding substantially lower values of $\rho$ throughout all other regions. Despite these qualitative differences, the KL divergence values in Table~\ref{tab:tau_sensitivity} again demonstrate that while absolute KL values vary significantly with the choice of weighting function, the relative rankings of model performance remain stable. RC consistently outperforms symplectic architectures across all weighting schemes, underscoring the robustness of our comparative framework.

\section{Discussion and conclusion}
The LD framework proved indispensable for evaluating NN performance, complementing traditional error-based metrics in Hamiltonian system forecasting. By quantifying phase-space “climate,” LDs revealed structural discrepancies invisible to conventional metrics, such as distorted homoclinic orbits and misaligned fixed points. Importantly, regions of large pointwise error do not necessarily indicate a breakdown of the underlying geometric organization. Small frequency shifts or slight spatial displacements of invariant structures can accumulate into significant local deviations over time, while still preserving the global topology of the phase space. In such cases, LD-based diagnostics distinguish between mere trajectory misalignment and genuine structural distortion.

Constructing PDFs weighted by LD values provided a robust statistical framework for comparing learned dynamics against reference solutions. Combined with information-theoretic measures such as the KL divergence, this approach enabled rigorous assessment of how faithfully machine learning models reproduce the geometric structures that govern long-term Hamiltonian dynamics. Unlike pointwise error metrics, which may obscure critical topological features, LD-based evaluation directly links prediction accuracy with the preservation of dynamically significant manifolds, offering deeper insight into model limitations and failure modes across phase space.

Symplectic models (SympNet, H\'enonNet, GHNN) possess architectural advantages due to their built-in invertibility, which guarantees efficient backward integration at the same computational cost as forward propagation. In contrast, RC required additional “warm-up” trajectories to initialize the reservoir state near target regions, introducing overhead despite its strong forecasting performance.

Our systematic comparison revealed system-dependent performance patterns tied to Hamiltonian structure. For the Duffing oscillator, characterized by a low-dimensional separable Hamiltonian, both symplectic networks and RC performed well, accurately resolving regular and near-homoclinic trajectories even with limited training data. This success reflects the alignment between the system’s mathematical structure and the inductive biases of these architectures.  

For the three-mode nonlinear Schr\"odinger equation (NLS), however, symplectic architectures underperformed. Although they preserved energy globally, GHNN and related models failed to capture the non-separable Hamiltonian’s phase-space topology, producing distorted homoclinic orbits. This gap highlights the limitations of current geometric priors when applied to more complex or chaotic Hamiltonian systems. RC, by contrast, achieved superior reconstruction of both short-term trajectories and global phase-space geometry, underscoring the value of architectural flexibility over strict symplecticity in such cases. Interestingly, GHNN’s difficulty in forecasting short-term dynamics may stem from frequency drift, suggesting that while the model captured global structures, it struggled with local temporal accuracy.

The observed underperformance of symplectic networks may reflect architectural or training limitations rather than inherent flaws in the symplectic approach. The GHNN tested here, for instance, used only two layers with 50 neurons each (Table~\ref{tab:nn_config_2mode}), likely insufficient to resolve the intricate couplings in a non-separable Hamiltonian. Deeper architectures, adaptive activation functions, or modified training strategies could potentially improve their ability to model chaotic regions.  

While the LD-framework is robust with respect to the integration time $\tau$, a comprehensive investigation of sensitivity to other key parameters remains essential for future work.  Of particular importance are the exponent $c$, the trajectory functional $\mathcal{M}$, the weighting function $g$ and the distance metric. These parameters fundamentally shape how dynamical structures are detected and quantified, as illustrated by the harmonic oscillator analysis in Appendix \ref{App:LD_Ham}. Several examples in the literature include using the action integral as the trajectory functional \cite{garcia-garrido2022LagrangianDescriptors} and weighting the function $g$ using second order derivative of LD rather than the value of LD itself \cite{daquin2022GlobalDynamics}. Such investigations would not only enhance the robustness of LD-based model evaluation but also establish parameter guidelines that adapt to specific system characteristics, strengthening the theoretical foundations of this approach while expanding its practical applicability to more complex Hamiltonian systems.

Future work should also address scalability. Extending the LD framework to quantify “climate fidelity” in higher-dimensional Hamiltonian systems, including higher-mode truncations of NLS dynamics, will be critical for benchmarking surrogate models. Dimensional-reduction methods, which have proven effective in dissipative systems such as the Kuramoto–Sivashinsky equation \cite{fleddermann2025ImprovingPrediction}, may be adapted to Hamiltonian settings. Our earlier investigations of full NLS dynamics using parallel RC \cite{hasmi2026ModelfreeForecasting} illustrate the challenges of building models that simultaneously preserve energy conservation and capture full dynamical complexity.

In the present work, Lagrangian Descriptors are employed exclusively as a post-processing diagnostic and do not enter the training procedure. An important direction for future research is to incorporate LD-based information directly into the learning process.

At least two possible approaches can be envisioned: (i) augmenting the loss function with a term that penalizes discrepancies between the LD fields of the learned and reference systems, thereby explicitly enforcing geometric consistency; and (ii) using LD-derived quantities as weighting factors within a standard reconstruction loss to emphasize dynamically relevant regions of phase space. While the first approach is conceptually appealing, it presents several practical challenges. Computing LD fields requires trajectory integrations and, depending on the formulation, additional variational calculations. Incorporating such quantities within the training loop would significantly increase computational cost. In contrast, the second approach—using LD-based weights within a conventional loss—may offer a more computationally tractable path toward geometry-informed training. 

In summary, LD-based analysis provides a principled framework for connecting local trajectory errors with global dynamical structures, enabling rigorous evaluation of machine-learning surrogates for Hamiltonian dynamics. Hybrid architectures that combine geometric priors with the flexibility of data-driven approaches appear especially promising for advancing reliable long-term simulations of complex Hamiltonian systems.

\section*{CRediT authorship contribution statement}
The manuscript was written with contributions from all authors. All authors have given their approval to the final version of the manuscript.\\

\textbf{Abrari Noor Hasmi:} Conceptualization, Software, Formal Analysis, Visualization, Writing - original draft, \textbf{Haralampos Hatzikirou:} Conceptualization, Supervision, Writing - review \& editing, \textbf{Hadi Susanto:} Conceptualization, Supervision, Writing - review \& editing.

\section*{Acknowledgment}
Khalifa University supported this work with a Faculty Start-Up Grant (No.\ 8474000351/FSU-2021-011). We acknowledge the contribution of Khalifa University's high-performance computing and research computing facilities in providing computational resources for this research. HS also acknowledges support by Khalifa University through a Research \& Innovation Grant under project ID KU-INT-RIG-2024-8474000789. The authors thanked the reviewer for their careful reading and the remarks that improved the manuscript.

\section*{Declaration of generative AI and AI-assisted technologies in the writing process}
During the preparation of this work, the authors used Grammarly and ChatGPT in order to improve language and readability. After using these tools/services, the authors reviewed and edited the content as needed and take full responsibility for the content of the publication.

\section*{Code Availability}
The GHNN architecture employed in this study is adapted from the publicly available implementation (\url{https://github.com/AELITTEN/NeuralNets_GHNN}).The RC models are based on the ReservoirComputing.jl package from the SciML project (\url{https://github.com/SciML/ReservoirComputing.jl}). The prototype code used to reproduce the numerical results of this study is publicly available at: \url{https://github.com/abrari-ku/LD_in_NN_Hamiltonian}
\appendix
\section{Lagrangian Descriptor field }
In this appendix, we provide the Lagrangian Descriptor (LD) fields corresponding to the LD-weighted PDF phase space shown in Fig.~5. These plots complement the LD-weighted probability density functions presented in the main text and allow for a direct qualitative comparison between the underlying geometric structures and their statistical representation.

\begin{figure}
    \centering
    \includegraphics[width=0.49\linewidth]{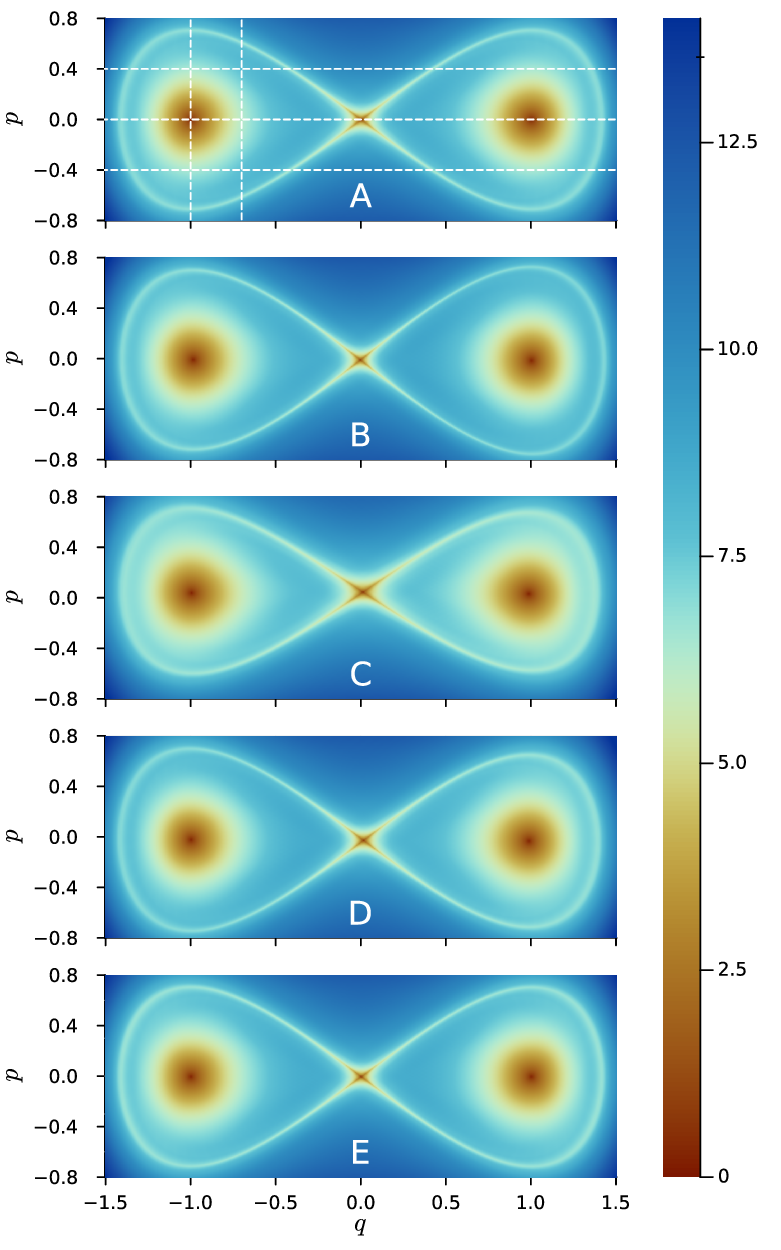}
    \includegraphics[width=0.49\linewidth]{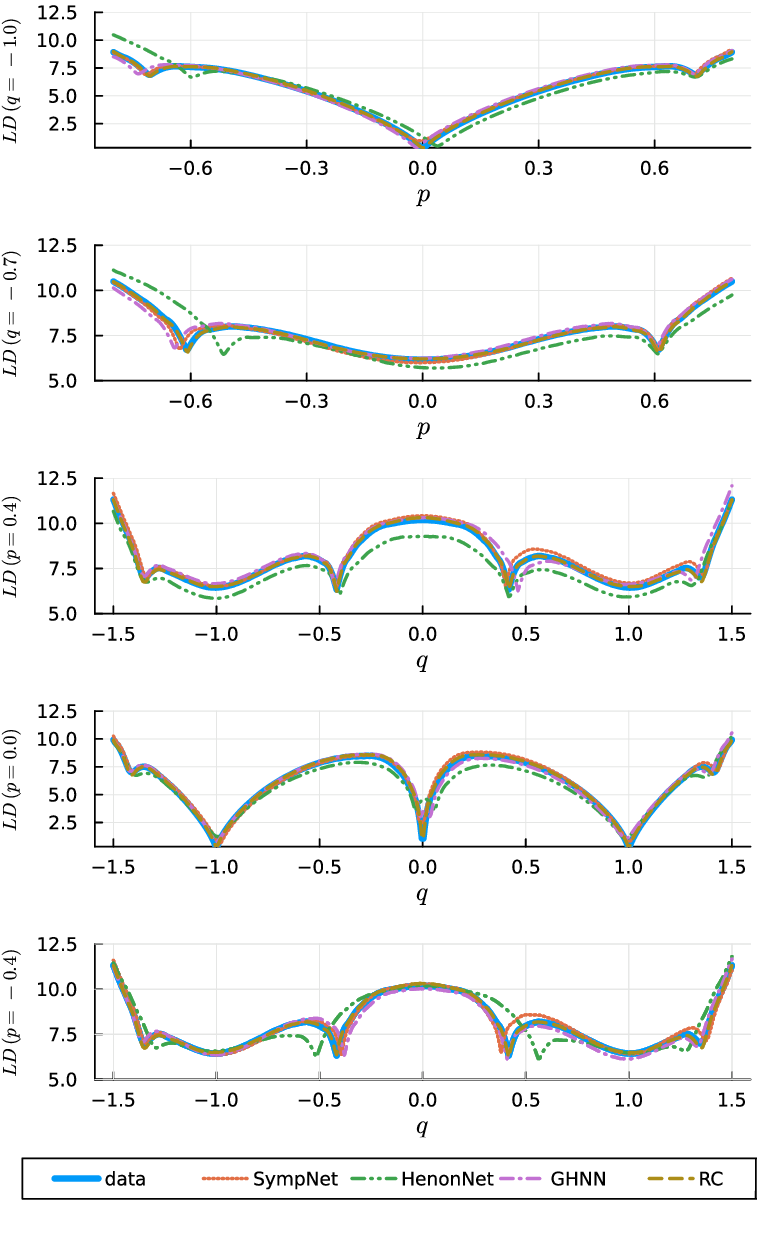}
    \caption[LD-weighted PDF for Duffing equation.]{LD for the Duffing equation correspond to Fig.~\ref{fig:LD_PDF_Duffing}. The left panels: (A) Numerical integration, (B) SympNet, (C) H\'enonNet, (D) GHNN, (E) RC. All NNs were trained on the dataset corresponding to the 200-trajectory case. High-density regions correspond to slow dynamics, revealing the fixed points near $(q,p)=(\pm1,0)$. ight panels: one-dimensional slices of the LD-weighted PDF across all methods at selected phase-space locations. 
From top to bottom: $q=-1.0$, $q=-0.7$, $p=-0.4$, $p=0.0$, and $p=0.4$.}
    \label{fig:LD_Duffing}
\end{figure}
\section{Illustrative Example of LD-weighted PDF\label{App:LD_Ham}}
\markright{Illustrative Example of LD-weighted PDF}
This appendix provides a relation between LD-weighted PDFs and the Hamiltonian in a simple system, namely the harmonic oscillator, where exact solutions are available. Most of these results are already derived in \cite{naik2019FindingNHIM}. Consider the harmonic oscillator with Hamiltonian:
\begin{equation}
    \mathcal{H} = \frac{\omega}{2}(p^2 +q^2).
\end{equation}
The equations of motion are given by:
\begin{align}
    \dot{q} &= \omega p, \\
    \dot{p} &= -\omega q.
\end{align}
Without loss of generality, we consider the following analytical solution:
\begin{align}
    q(t) &= -A \cos(\omega t) \\
    p(t) &= A \sin(\omega t),
\end{align}
where \(A = \sqrt{\frac{2H}{\omega}}\) and H is the system's energy, which depends on the initial conditions. 
To calculate the LD, we use the property that $|\sin(\omega t)|$ and $|\cos(\omega t)|$ are periodic functions with period \(T = \frac{\pi}{\omega}\). Hence:
\begin{align}
    \int_0^{\frac{\pi}{2\omega}} |\dot{q}|^c dt &= |A\omega|^{c}\int_0^{\frac{\pi}{2\omega}} | \sin(\omega t)|^c dt \\ 
    &= \frac{(2\omega H)^{c/2}}{2\omega}B\left(\frac{1}{2},\frac{c+1}{2}\right) = \frac{\sqrt{\pi}(2\omega H)^{c/2}}{c\omega}\frac{\Gamma\left(\frac{c+1}{2}\right)}{\Gamma\left(\frac{c}{2}\right)},
\end{align}
where $B$ and $\Gamma$ are the Beta and Gamma functions, respectively. A similar calculation applies for \(|\dot{p}|\). Let $\tau = NT + r$ where $N$ is an integer and $0\leq r<T$ is the remainder term. The LD is given by: 
\begin{align}
    LD(u_0) &= \int_{-\tau}^{\tau}|\dot{q}|^c +|\dot{p}|^c dt,\\
    & = \frac{8N\sqrt{\pi}(2\omega H)^{c/2}}{c\omega} \frac{\Gamma\left(\frac{c+1}{2}\right)}{\Gamma\left(\frac{c}{2}\right)} + \int_{-r}^{r}|\dot{q}|^c +|\dot{p}|^c dt,\\
    & = \frac{8(2\omega)^{c/2}(\tau-r)}{c\sqrt{\pi}} \frac{\Gamma\left(\frac{c+1}{2}\right)}{\Gamma\left(\frac{c}{2}\right)} H^{c/2}+ \int_{-r}^{r}|\dot{q}|^c +|\dot{p}|^c dt.
\end{align} 
For large integration times $\tau$, the LD is dominated by the first term, with the asymptotic value given by: 
\begin{equation}
\lim_{\tau\to\infty}\frac{LD(u_0)}{\tau}  = \frac{8(2\omega)^{c/2}}{c\sqrt{\pi}}\frac{\Gamma\left(\frac{c+1}{2}\right)}{\Gamma\left(\frac{c}{2}\right)}H^{c/2},
\end{equation}
where the only spatial dependence is in the energy \(H\). The LD-weighted PDF should be interpreted as a normalization of $g(H^{c/2})$ by its integral over the domain. This demonstrates how the LD-weighted PDF naturally highlights structures of different energies in the phase space.

\section{Polar form of the three-mode NLS}\label{App:PolarForm3Mode}
Substituting Eq.~\eqref{eq:polar_subs} into Eqs.~\eqref{eq:2mode_q0}–\eqref{eq:2mode_p1}, we obtain
\begin{align}
i\dot{\zeta}_0 - \zeta_0\dot{\theta}_{0} + \zeta_0^{3} + \zeta_0\zeta_1^{2}e^{2i(\theta_{1}-\theta_{0})} + 2\zeta_0\zeta_1^{2} &= 0,\\
i\dot{\zeta}_0 + \zeta_0\dot{\theta}_{0} - \zeta_0^{3} - \zeta_0\zeta_1^{2}e^{-2i(\theta_{1}-\theta_{0})} - 2\zeta_0\zeta_1^{2} &= 0,\\
i\dot{\zeta}_1 - \zeta_1\dot{\theta}_{1} - \tfrac{1}{2}k^{2}\zeta_1 + \tfrac{3}{2}\zeta_1^{3} + \zeta_0^{2}\zeta_1e^{-2i(\theta_{1}-\theta_{0})} + 2\zeta_0^{2}\zeta_1 &= 0,\\
i\dot{\zeta}_1 + \zeta_1\dot{\theta}_{1} + \tfrac{1}{2}k^{2}\zeta_1 - \tfrac{3}{2}\zeta_1^{3} - \zeta_0^{2}\zeta_1e^{2i(\theta_{1}-\theta_{0})} - 2\zeta_0^{2}\zeta_1 &= 0.
\end{align}
Defining the phase difference \(\phi = \theta_1-\theta_0\) and simplifying, the system reduces to
\begin{align}
\dot{\zeta}_0 &= -\zeta_0\zeta_1^{2}\sin(2\phi), \\
\dot{\zeta}_1 &= \ \ \zeta_0^{2}\zeta_1\sin(2\phi), \\
\dot{\phi} &= \tfrac{1}{2}k^{2} + \zeta_0^{2} - \tfrac{1}{2}\zeta_1^{2} + (\zeta_0^{2}-\zeta_1^{2})\cos(2\phi).
\end{align}
Next, introduce the normalized variable \(\eta = \zeta_0^{2}/P_{0}\), and rescale time and wavenumber as \(\hat{t} = P_{0}t\) and \(\hat{k} = k\sqrt{P_{0}}\). Dropping hats for simplicity, the equations become
\begin{align}
\dot{\eta} &= -2\eta(1-\eta)\sin(2\phi),\\
\dot{\phi} &= \tfrac{1}{2}(k^{2}-1) + \tfrac{3}{2}\eta - (1-2\eta)\cos(2\phi),
\end{align}
which correspond to a Hamiltonian system with
\begin{equation}
\mathcal{H}_{3p} = \eta(1-\eta)\cos(2\phi) + \tfrac{1}{2}(1-k^2)\eta - \tfrac{3}{4}\eta^{2}.
\end{equation}
The homoclinic orbit is given by the level set $\mathcal{H}_{3p}=0$, yielding Eq.~\eqref{eq:Hom_2D}

\section{Computational cost of NN models} \label{App:CompCost}
This appendix provides a detailed analysis of the computational complexity of the NN architectures considered in this work. We establish theoretical bounds on algorithmic efficiency using FLOPs as the primary measure.

Our analysis focuses on inference-time complexity rather than training costs, since inference efficiency directly impacts the practical utility of these models in real-time forecasting applications. For tractability, we adopt the following simplifying assumptions:
\begin{itemize}
    \item Memory access patterns, cache effects, and hardware-specific optimizations are neglected;
    \item Each activation function evaluation is counted as one FLOP per element;
    \item All models are assumed to use the same numerical precision (e.g., float32).
\end{itemize}

While these assumptions omit certain implementation-specific details, they provide a consistent and hardware-agnostic basis for comparing the intrinsic computational efficiency of different architectures. The reported FLOP counts should therefore be interpreted as asymptotic complexity estimates rather than precise runtime measurements.

\subsection*{SympNets}
For SympNets, we analyze the cost of a single gradient module of the form
\begin{equation}
 \nabla V(q) = K^\top \,\text{diag}(a)\, \sigma(Kq+b),
\end{equation}
where $K \in \mathbb{R}^{m\times n}$, $a \in \mathbb{R}^m$, and $\sigma$ is the activation function. The FLOP breakdown is:
\begin{itemize}
    \item Matrix–vector multiplication $Kq$: $2mn - m$
    \item Vector addition $Kq + b$: $m$
    \item Activation $\sigma(Kq+b)$: $m$
    \item Elementwise multiplication $\text{diag}(a)\,\sigma(\cdot)$: $m$
    \item Matrix–vector multiplication $K^\top(\cdot)$: $2mn - n$
    \item Addition with momentum term $\nabla V(q) + p$: $n$
\end{itemize}
Total FLOPs per gradient module: $2m(2n+1)$.  

Each Hamiltonian layer requires two gradient evaluations (for $q$ and $p$), giving
\[
\text{FLOPs per layer} = 2(4mn+2m) = 8mn+4m.
\]
For $l$ learned Hamiltonians,
\begin{equation}
    \text{FLOPs}_{\text{SympNet}} = l(8mn+4m) = 4lm(2n+1).
\end{equation}

\subsection*{H\'enonNets}
For a H\'enonNet with one hidden layer, the gradient has a structure similar to Eq.~\eqref{eq:V}. The cost of a single H\'enon map is
\begin{equation}
    \text{FLOPs}_{\text{H\'enon map}} = 4mn + 2m + n.
\end{equation}
A H\'enon layer consists of four such maps:
\begin{equation}
    \text{FLOPs}_{\text{H\'enon layer}} = 4(4mn+2m+n).
\end{equation}
Since each H\'enon layer learns two Hamiltonians, the total cost for $l$ learned Hamiltonians is
\begin{equation}
    \text{FLOPs}_{\text{H\'enonNet}} = 2l(4mn+2m+n) = 8lmn + 4lm + 2ln.
\end{equation}

\subsection*{Generalized Hamiltonian Neural Networks (GHNN)}
GHNNs generalize both SympNets and H\'enonNets. Here we consider a GHNN with two hidden layers, trained using a symplectic Euler update per layer. The gradient takes the form
\begin{equation}
\nabla V(q) = K^\top \,\text{diag}\!\left(K_2^\top \text{diag}(a)\,\sigma\!\big(K_2\Sigma(Kq+b)+b_2\big)\right)\sigma(Kq+b).
\end{equation}
The FLOP breakdown is:
\begin{itemize}
    \item First affine transform $Kq+b$: $2mn+m$
    \item First activation $\Sigma(Kq+b)$: $m$
    \item Second affine transform $K_2\Sigma(\cdot)+b_2$: $2m^2+m$
    \item Second activation $\sigma(\cdot)$: $m$
    \item Elementwise multiplication with $\text{diag}(a)$: $m$
    \item Matrix–vector multiplication $K_2^\top(\cdot)$: $2m^2-m$
    \item Reuse $\sigma(Kq+b)$: $m$
    \item Elementwise product of two activation streams: $m$
    \item Final multiplication $K^\top(\cdot)$: $2mn-n$
    \item Addition with momentum $+p$: $n$
\end{itemize}
Total FLOPs per (two-layer) gradient module: $4m(m+n+1)$.  

Since each learned Hamiltonian requires two gradient evaluations (for $q$ and $p$), the total cost is
\begin{equation}
    \text{FLOPs}_{\text{GHNN}} = 8lm(m+n+1).
\end{equation}

\subsection*{Reservoir Computing (RC)}
We now analyze the computational cost of inference in the reservoir computing model, defined as
\begin{align}
    x_{k+1} &= (1-\alpha)\,x_k + \alpha\,\tanh(W_x x_k + W_u u_k),\\
    y_{k+1} &= W x_{k+1},
\end{align}
where $u_k \in \mathbb{R}^{N_u}$ is the input, $x_k \in \mathbb{R}^{N_x}$ is the reservoir state, and $y_{k+1} \in \mathbb{R}^{N_u}$ is the output. The matrices $W_x \in \mathbb{R}^{N_x\times N_x}$, $W_u \in \mathbb{R}^{N_x\times N_u}$, and $W \in \mathbb{R}^{N_u\times N_x}$ are the reservoir, input, and output weight matrices, respectively. The parameter $\alpha$ denotes the leak rate.

The FLOP breakdown for one inference step is:
\begin{itemize}
    \item Reservoir update $W_x x_k$: $2N_x^2 - N_x$
    \item Input update $W_u u_k$: $2N_x N_u - N_x$
    \item Nonlinear activation and scaling $\alpha\tanh(\cdot)$: $3N_x$
    \item Leak update $(1-\alpha)x_k + \alpha(\cdot)$: $2N_x$
    \item Output readout $W x_{k+1}$: $2N_u N_x - N_u$
\end{itemize}
Hence, the total cost per forecasting step is
\begin{equation}
    \text{FLOPs}_{\text{RC}} = 2N_x^2 + 4N_x N_u + 3N_x - N_u.
\end{equation}

The above estimate assumes a dense reservoir matrix $W_x$. In practice, $W_x$ is typically sparse, which reduces the cost significantly. Let $\mu$ denote the fraction of nonzero entries in $W_x$. Then the FLOP count becomes
\begin{equation}
    \text{FLOPs}_{\text{RC}} = 2\mu N_x^2 + 4N_x N_u + 3N_x - N_u.
\end{equation}

These estimates apply to each inference step. Additional costs associated with reservoir warm-up to initialize the state are excluded here.

\bibliographystyle{elsarticle-num} 
\bibliography{Reference.bib}

\end{document}